\documentclass[final,5p,times,twocolumn]{elsarticle}

\usepackage{amsmath,amscd,amssymb}
\usepackage{bm}
\usepackage{graphicx}
\usepackage[tight]{subfigure}
\usepackage{units}
\usepackage{array,booktabs}
\usepackage{amssymb}
\usepackage{setspace}
\usepackage{color}

\journal{Ultrasonics}

\begin{document}

\begin{frontmatter}

\title{Deconvolution of vibroacoustic images using a simulation model based on a three dimensional point spread function}
\author[1]{Talita Perciano\fnref{nota1}}
\fntext[nota1]{Presently associated with Instituto de Matem\'atica e Estat\'istica, 
Universidade de S\~ao Paulo, S\~ao Paulo, SP 05508-090, Brazil}
\author[2]{Matthew W. Urban}
\author[1]{Nelson D. A. Mascarenhas}
\address[1]{Departamento de Computa\c c\~ao, Universidade Federal de S\~ao Carlos, 
S\~ao Carlos, SP 13565-905, Brazil.}
\author[2]{Mostafa Fatemi}
\address[2]{Department of Physiology and  Biomedical Engineering, Mayo Clinic College of Medicine, Rochester, MN 55905, United States}
\author[3]{Alejandro C.\ Frery}
\address[3]{Instituto de Computa\c c\~ao, Universidade Federal de Alagoas, Macei\'o, AL 57072-970, Brazil}
\author[4]{Glauber T.\ Silva\corref{cor1}}
\cortext[cor1]{Corresponding author: \texttt{glauber@pq.cnpq.br}}
\address[4]{Physical Acoustics Group, Instituto de F\'isica, Universidade Federal de Alagoas, Macei\'o, AL 57072-970, Brazil}

\begin{abstract}
Vibro-acoustography (VA) is a medical imaging method based on the dif\-ference-frequency generation produced by 
the mixture of two focused ultrasound beams.
VA has been applied to different problems in medical imaging such as
imaging bones, microcalcifications in the breast, mass lesions, and calcified arteries. 
The obtained images may have a resolution of $0.7$--$\unit[0.8]{mm}$.
Current VA systems based on confocal or linear array transducers generate C-scan images at the beam focal plane.
Images on the axial plane are also possible, however the system resolution along depth worsens when
compared to the lateral one. Typical axial resolution is about $\unit[1.0]{cm}$.
Furthermore, the elevation resolution of linear array systems is larger than that in lateral direction.
This asymmetry degrades C-scan images obtained using linear arrays. 
The purpose of this article is to study VA image restoration based on a 3D point spread function (PSF)
using classical deconvolution algorithms: Wiener, constrained least-squares (CLSs), and geometric mean filters.
To assess the filters' performance on the restored images, we use an image quality index that accounts for correlation 
loss, luminance and contrast distortion.
Results for simulated VA images show that the quality index achieved with the Wiener filter is $0.9$ (when the index 
is $1$ this indicates perfect restoration).
This filter yielded the best result in comparison with the other ones.
Moreover, the deconvolution algorithms were applied to an experimental VA image of a phantom
composed of three stretched $\unit[0.5]{mm}$ wires.
Experiments were performed using transducer driven at two frequencies, 
$\unit[3075]{kHz}$ and $\unit[3125]{kHz}$, which resulted in the difference-frequency of $\unit[50]{kHz}$.
Restorations with the theoretical line spread function (LSF) did not recover sufficient information 
to identify the wires in the images.
However, using an estimated LSF the obtained results displayed enough information to spot the wires in the images.
It is demonstrated that the phase of the theoretical and the experimental PSFs are dissimilar.
This fact prevents VA image restoration with the current theoretical PSF.
This study is a preliminary step towards understanding the restoration of VA images through 
the application of deconvolution filters.
\end{abstract}


\end{frontmatter}


\section{Introduction}
Recent practices of medicine strongly rely on the visualization of inner organs
in order to study the relationship of anatomic structure and physiological process 
of living beings.
In this context, vibro-acoustography (VA)~\cite{fatemi98} has become a medical imaging tool for investigating 
calcifications in human arteries~\cite{greenleaf98,fatemi00,fatemi03}, 
microcalcifications in the breast~\cite{fatemi01,wold04}, liver tumors~\cite{alizad04}, 
bone structure~\cite{calle01}, among others applications~\cite{urban:2011}.

\begin{figure}[htb!]
\centering
 \includegraphics[width=\linewidth]{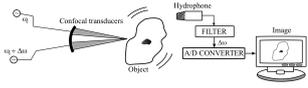}
 \caption{Description of a typical VA imaging system.
 The central- and the difference-frequency are $\omega_0$ and $\Delta \omega$, respectively.}
 \label{fig:figure1}
\end{figure}

The VA technique employs two co-focused ultrasound beams in megahertz range with slightly different 
frequencies, typically in the kilohertz range, to scan biological tissues 
(see Fig.~\ref{fig:figure1} for the system description).
The VA image formation process is related to the nonlinear interaction of the primary ultrasound waves (incident and scattered),
which gives rise to a secondary acoustic wave at the difference-frequency~\cite{silva:1326,silva:2011}.  
The generated difference-frequency wave is detected by a hydrophone (typical bandwidth $1$--$\unit[70]{kHz}$).
The acquired signal is processed and used to form VA images.

The interaction of the primary incident ultrasound waves gives rise to a highly collimated difference-frequency beam, 
known as the parametric array effect~\cite{westervelt:535}.
It has been recently shown that the difference-frequency signal
detected in the forward scattering direction is mostly due to the nonlinear interaction of the primary scattered 
waves~\cite{silva:2011}. 
Hence, the parametric array may behave as a background
signal present in the difference-frequency wave detected to form VA images.

The VA method can be extended to generate several images at once using a
multifrequency arrangement also known as multifrequency VA~\cite{urban06}. 
In this configuration, $N$ ultrasound beams (more than two) are produced and
they overlap in the system focal region. 
Due to the nonlinear interaction of the ultrasound waves,
$N(N-1)/2$ difference-frequency components arise in the medium and they are used
to form multiple images with just one scanning process.

The spatial resolution of VA described in terms of the system point spread function (PSF)
is determined by the transducer geometrical properties and the ultrasound driving frequencies.
Beamforming has a great impact on the system resolution.
Some schemes to generate two or more ultrasound beams for VA  have been proposed based on a confocal transducer~\cite{fatemi99}, 
$x$-focal (crossing beams)~\cite{chen04},
sector array~\cite{glauber04}, linear array~\cite{glauber04,urban11}, and reconfigurable array~\cite{kamimura:}. 
The VA resolution does not to correspond to the difference-frequency beam waist 
due to the parametric array effect. 
Indeed, the system PSF has a narrower lateral width than that of a parametric beam produced by the VA confocal transducer.
To verify this, we have computed the parametric beam with the theory presented in Ref.~\cite{ding:2759}
and found that this beam does not fit the PSF waist shown in Ref.~\cite{fatemi99}. 
However, for the sake of brevity, these results will not be included in this article.

The study presented here is limited to VA systems based on the confocal arrangement.
Specifically, a confocal VA system composed by a two-element confocal transducer
with $\unit[44]{mm}$-diameter, focused at $\unit[70]{mm}$, and
operating with frequencies around $\unit[3.1]{MHz}$, achieves a spatial resolution 
($\unit[-6]{dB}$ beamwidth)
of approximately \unit[$0.8\times 0.8$]{mm} in the focal plane and \unit[$16$]{mm} axially.
With this resolution only images formed at the focal plane are feasible.
Furthermore, compared to conventional pulse-echo ultrasound images, VA images generated by the confocal transducer do 
not exhibit speckle noise.

Clinical applications of VA will demand systems based on linear array transducers because the ultrasound beams
can be electronically steered over the region to be imaged.
It has been shown through computational simulations that a typical linear array 
with $\unit[32]{mm}$-aperture operating at $\unit[3]{MHz}$ can produce 
VA images with $\unit[-6]{dB}$ of lateral resolution of $\unit[1.9]{mm}$ when 
focused at $\unit[5]{cm}$ in depth~\cite{glauber04}.
However, the system PSF is not symmetric in the imaging plane as in other VA beamforming schemes.
The resolution is poorer in the elevation when compared to the lateral direction (roughly threefold larger).
Hence, linear array systems (1.25 and 1.5D) may produce distorted VA images.
Recently, a VA system based on the General Electric Vivid 7 ultrasound scanner with the GE 7L and 10L
linear array transducers was designed and assessed~\cite{urban11}. This system was set to operate at 
nearly $\unit[5]{MHz}$, with the difference-frequency being $\unit[51.54]{kHz}$.
When the incident beams were focused at $\unit[25]{mm}$ in depth, the spatial resolution  
achieved by the system at $\unit[-6]{dB}$ was
$\unit[0.573]{mm}$ laterally, $\unit[1.44]{mm}$ in elevation,  and $\unit[2.20]{mm}$ axially.

Limitations in the spatial resolution of VA systems degrade the yielded images, reducing
their utility to medical imaging applications.
The need for restoration of VA images in either elevation (with linear arrays) or axial directions
has prompted us to assess the efficiency of some restoration algorithms applied to these images.
In doing so, three deconvolution frequency domain filters, namely constrained least squares (CLS), 
geometric mean, and Wiener filters~\cite{gonzalezcap5,hunt73} based on the theoretical PSF of the system are 
employed to restore simulated VA images~\cite{perciano08}.
In addition, a phantom composed by three stretched $\unit[0.5]{mm}$ wires was imaged by a confocal VA system.
The VA line spread function (LSF) was estimated from the image and used to restore the image through the CLS filter. 
Restoration with the theoretical LSF yielded results with no visual information of the wires. 
The main reason for this problem lies on the lack of the phase information in the theoretical LSF.
This is illustrated by restoring the actual VA image with a combined LSF, in which the magnitude and the phase come from 
the theoretical and the experimental LSF, respectively.
In this case, the CLS filter restoration is capable to restore most of the information such as position, shape and contrast
of the wires in the image.

This article is organized as follows.
A model for  vibro-acoustic image formation based on a three-dimensional PSF composed by the product 
of the pressure amplitudes of the incident beams is presented in Sec.~\ref{sec:va}. 
The general aspects of the restoration filters (Wiener, CLS, and geometric mean) are briefly described in 
Sec.~\ref{sec:restoration}.
The computational and experimental results are explained in Sec.~\ref{sec:results}.  
In Sec.~\ref{sec:conclusions}, the main conclusions of this study are outlined. 

\section{Image formation}
\label{sec:va}
The VA technique employs two ultrasound beams at frequencies $\omega_1$ and $\omega_2$ to form
an image of the region-of-interest.
Hence the primary incident pressure upon this region is given by
$$
p(\mathbf{r},t) = p_1(\mathbf{r}) e^{-i\omega_1 t} + p_2(\mathbf{r})
e^{-i\omega_2 t},
$$
where $\mathbf{r}$ is the position vector indicating the observation point, $i$ is the imaginary unit,
$t$ is the time, 
and $p_1$ and $p_2$ are the complex pressure amplitudes.

Based on the radiation force concept, a 2D image formation model was proposed in which
the system PSF is the product of the incident pressure fields in the transducer
focal plane~\cite{fatemi99}.
An extension of this model to account for depth-of-field effects in the VA image formation was presented in 
Ref.~\cite{glauber06}. 
According to these image formation models, the three-dimensional PSF of a VA system is given by
\begin{equation}
\label{psf1}
 h(\mathbf{r})  = p_1^*(\mathbf{r})p_2(\mathbf{r}),
\end{equation}
where the symbol * means complex conjugation.
It is worthy noting that the PSF of a VA system as given in Eq.~(\ref{psf1}) is a complex function.
Despite early success on explaining VA as a radiation force imaging method,
a recent study~\cite{silva:2011} has shown that the difference-frequency signal used to form VA images
depends on the nonlinear interaction of primary incident and scattered waves at the frequencies $\omega_1$ and $\omega_2$.
Both radiation force phenomenon and the nonlinear interaction of the waves scale quadratically with the primary
pressure amplitudes.
Hence, the description of the difference-frequency pressure magnitude based upon the radiation force and the nonlinear
interaction models are alike, but the phase information due to each phenomenon is different~\cite{silva:2011}.
Therefore, as the deconvolution of VA images needs to be done with the PSF (magnitude and phase).
Thus, we can expect that the deconvolution using the PSF given in Eq.~(\ref{psf1}) may not be able
to fully recover the visual information in VA images.

\begin{figure}[hbt!]
 \centering
 \mbox{\subfigure{\label{fig:figure2}\includegraphics[width = 0.7\linewidth,angle=-90]{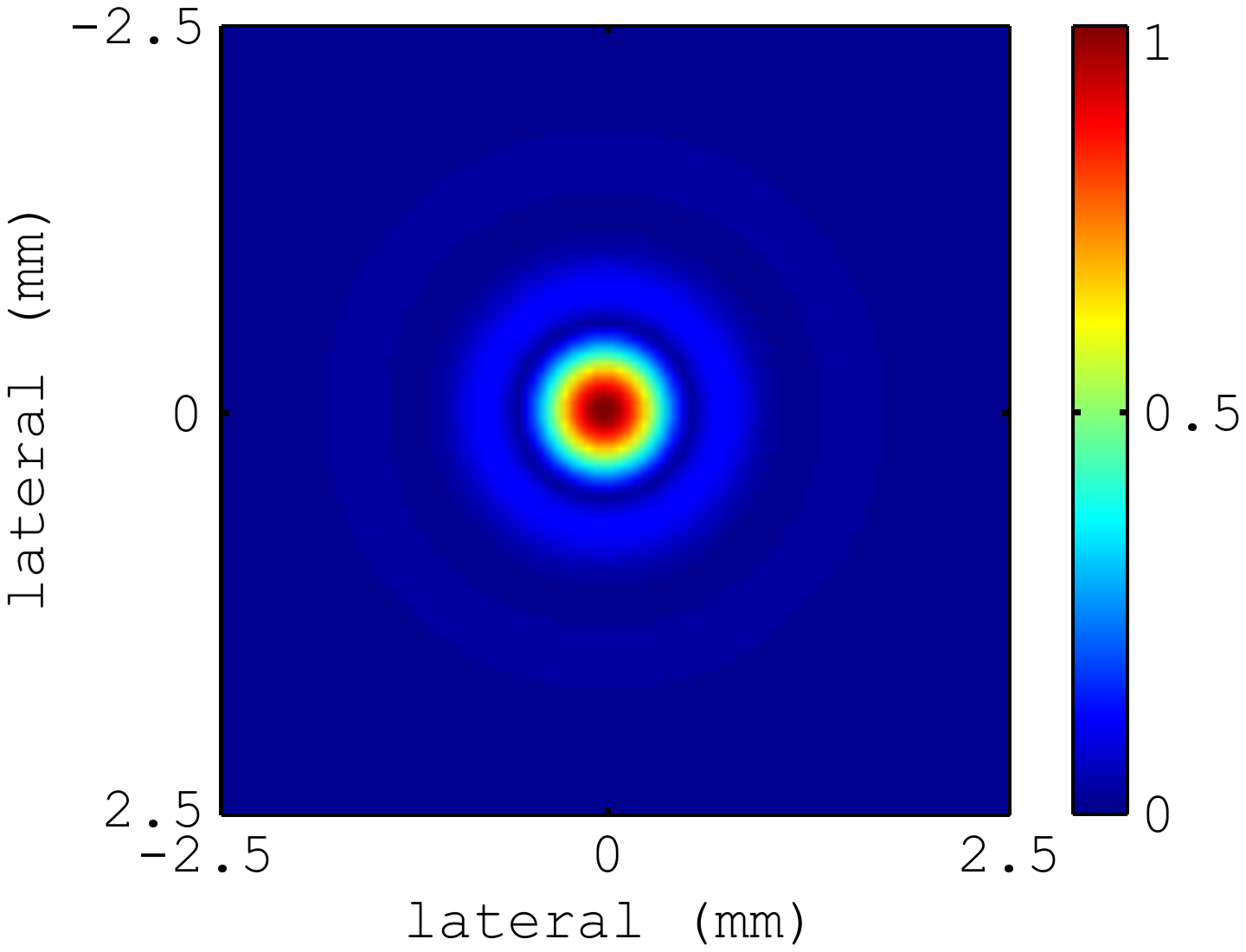}}}
 \mbox{\subfigure{\label{fig:figure3}\includegraphics[width = 0.44\linewidth,angle=-90]{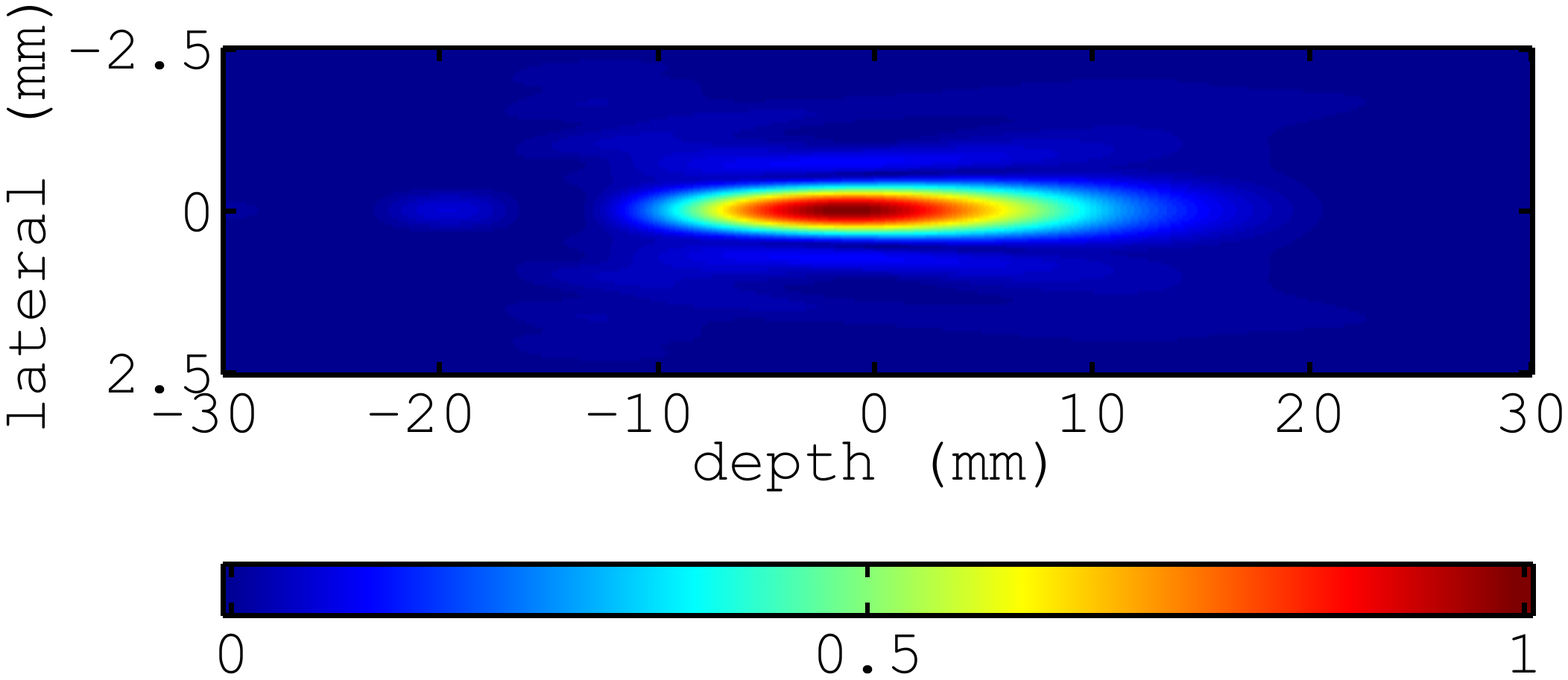}}}
\caption{Magnitude of the PSF in the focal plane and in depth for a two-element confocal transducer 
with inner and outer radii being respectively $14.8$ and $\unit[22]{mm}$. 
The transducer focal distance is $\unit[70]{mm}$, which corresponds to zero in the depth scale.}
\label{fig:psf}
\end{figure}

To compute the PSF given in Eq.~(\ref{psf1}) one has
to calculate the incident pressure field amplitudes $p_1$ and $p_2$ in the linear approximation for the VA transducer.
This has been done using the Field II simulation program~\cite{jensen:262}.
Even though the fields from Field II are linear the combination through equation 2 makes the problem inherently nonlinear. 
The nonlinearity of the VA beamforming arises because of the interaction of the two beams, 
not the nonlinear propagation of the pressure from the transducer.
The simulations were performed considering
the transducer and operation frequencies used in the experiments to be described later.
Hence, a two-element spherical confocal transducer with a
$\unit[44]{mm}$ diameter and focus 
at $\unit[70]{mm}$ was considered.
The transducer inner radius is $\unit[14.8]{mm}$, while the outer ring ranges
from $15.2$ to $\unit[22]{mm}$.
The transducer elements were driven independently by sinusoidal signals
at $\unit[3.075]{MHz}$ and $\unit[3.125]{MHz}$.
The magnitudes of the simulated PSF laterally and in depth are presented in Fig.~\ref{fig:psf}.

The VA image of an object can be obtained by convolving the function which represents the object with the 
system PSF. 
Accordingly, the image function is given by~\cite{glauber06}
\begin{equation}
g(\mathbf{r})=f(\mathbf{r})\star h(\mathbf{r}) +
n(\mathbf{r}),
\label{eq1}
\end{equation}
where $f$ is the function that represents the object, $n$ represents the additive Gaussian noise, and
the symbol $\star$ denotes the spatial convolution.
The additive Gaussian noise is the only source of noise present in VA images
due to thermal fluctuations present in the electronic devices of the VA system.
The noise level of VA images has been assessed.
Thus, a white noise with 
$\unit[20]{dB}$ signal-to-noise ratio (SNR) was chosen to be added to all simulated images here.
This level was considered by visual inspection of the VA simulated and actual images.

In the current arrangement of VA, real-time images cannot be done because the system requires 
four or five difference-frequency cycles to form a pixel. 
At $\unit[50]{kHz}$, this corresponds to $\unit[80]{\mu s}$ per pixel. 
To form images with $100 \times 100$ pixels and show them in $24$ frames per
second, one needs $\unit[4.1]{\mu s}$ for each pixel.

\section{Restoration algorithms}
\label{sec:restoration}

Among different filtering schemes to perform image deconvolution, we choose the Wiener, the  constrained least-squares (CLSs), and the geometric mean (GM) filters~\cite{gonzalezcap5}. 
This choice is based on the following digital image processing aspects.
Firstly, the chosen filters operate in the frequency domain, which is
computationally less expensive than to perform the filtering in the spatial domain. 
Secondly, inasmuch as the VA image formation is based on a spatial
invariant model with additive noise as given in Eq.~(\ref{eq1}),
one can use inverse filter methods for deconvolution.
In this scheme, the Wiener filter removes the image degradation without division by small values, which can lead to numerical overflow. 
However, this filter reduces the noise level by smoothing the image without sharpening the edges.
On the other hand, the CLS filter reduces the image degradation
and it allows the control of sharpening the edges through a regularization parameter. 
The CLS and the Wiener filters were used for the deconvolution of 
ultrasound images~\cite{yeoh2006}.
The GM filter combines the features of the Wiener and the CLS filters depending on the adjustment of two parameters.
The criteria used here to set the filters parameters will be discussed later.

To  describe the Wiener, the CLS, and the GM filters,
the image formation model in Eq.~(\ref{eq1}) is written in
the frequency domain as
\begin{equation}
 G({\bf k}) = F({\bf k})H({\bf k}) + N({\bf k}),
\label{eq3}
\end{equation}
where  ${\bf k}$ is the reciprocal of ${\bf r}$, and $F$, $G$, $H$, and $N$ are the Fourier transforms of the functions $f$,
$g$, $h$, and $n$, respectively.

\subsection{Wiener filter}

The Wiener filter is designed to minimize the mean squared 
error between the filtered and the ideal images.
In the frequency domain, this filter is given by~\cite{gonzalezcap5}
\begin{equation}
F = \left[\frac{H^*}{|H|^2 + S_n/S_f}\right]G,
\end{equation}
where $S_n$ and $S_f$ are the power spectrum density of the noise and of the ideal image, respectively.

\subsection{Constrained least-squares filter}
In the frequency domain, the constrained least-squares
(CLSs) filter is expressed as~\cite{gonzalezcap5}
\begin{eqnarray}
F = \left[\frac{H^*}{|H|^2+\gamma|P|^2}\right]G,
\end{eqnarray}
where $\gamma$ is the regularization parameter and $P$ is the Fourier Transform of the discrete Laplacian operator in three-dimensions.

\subsection{Geometric mean filter}
The geometric mean (GM) filter is given in the frequency
domain by~\cite{gonzalezcap5}
\begin{eqnarray}
F = 
\left\{\left[\frac{H^*}{|H|^2}\right]^\alpha\left[\frac{H^*}{|H|^2+\gamma
S_n/S_f}\right]^{1-\alpha}\right\}G,
\end{eqnarray}
where $\alpha \in [0, 1]$ and $\gamma>0$.

\subsection{Filter parameters estimation}
\label{subsec:parameters}
In performing the VA image deconvolutions, a trial and error 
approach is adopted.
In doing so, the best visual and quantitative results 
for the restored images are sought.
The quantitative analysis is performed using three indices of digital image processing, 
namely~\cite{wang02,WangBovikLu:ICASSP:02} improvement in the signal-to-noise ratio (ISNR), 
mean squared error (MSE), and universal image quality index (UIQI).
For the last two indices, the imaginary part of the restored image is ignored,
because  it approaches to zero after the filtering process.

\section{Results and discussion}
\label{sec:results}

\subsection{Computational simulations}
\label{subsec:simulations}

\begin{figure}[hbt!]
 \centering
 \includegraphics[width=\linewidth]{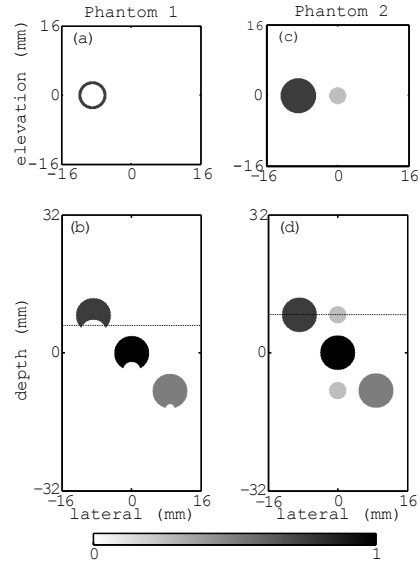}
 \caption{Phantoms used in the computational restoration. 
(a) Phantom 1 seen in the \unit[6]{mm}-transverse plane.
(b) Phantom 1 seen in-depth with three inclusions with \unit[$4$]{mm}
 radius each and  $\unit[36]{dB}$ (left),
$\unit[0]{dB}$ (center), and $\unit[42]{dB}$ (right) contrasts.
(c) Phantom 2 seen in the \unit[8]{mm}-transverse plane.
(d) Phantom 1 seen in-depth with two additional inclusions
of \unit[$2$]{mm} radius and $\unit[46]{dB}$ contrast.
The dotted lines in (b) and (d) indicates the position of the images (a) and (b).
}
 \label{fig:figure4}
\end{figure}

To perform the computational simulation of VA images, the PSF is represented by a 
complex matrix whose dimensions are $256\times 256\times 512$.
This corresponds to a region of \unit[$32$]{mm} laterally and \unit[$64$]{mm} in depth.
Two computational phantoms are devised with three and five spherical inclusions
as shown in Fig.~\ref{fig:figure4}.
The phantoms have dynamic range of $\unit[48]{dB}$, i.e., a pixel is
represented with $\unit[8]{bits}$ depth.
Later, we present two computer phantoms with inclusions having up to four contrast levels.
The purpose of using such phantoms is to demonstrate that the deconvolution of VA images can recover most information regarding to 
the inclusions morphology.
Thus, the adopted dynamic range, which represents inclusions with $256:1$ contrast ratio, is sufficient for this proposal. 
In turn, the inclusions can be regarded as soft tissue with different mechanical properties compared to the background.

Phantom 1 is composed by three inclusions having same radius of  \unit[$4$]{mm}.
The inclusions have contrasts of $\unit[36]{dB}$ (left),
$\unit[0]{dB}$ (center),  and $\unit[42]{dB}$ (right).
Phantom 2 is similar to phantom 1 with two additional spheres are included with \unit[$2$]{mm} radius each and $\unit[46]{dB}$
contrast.
The dimensions of the inclusions are typically found in VA images of mass legions and calcifications.

\begin{figure}[hbt!]
 \centering
 \includegraphics[width=\linewidth]{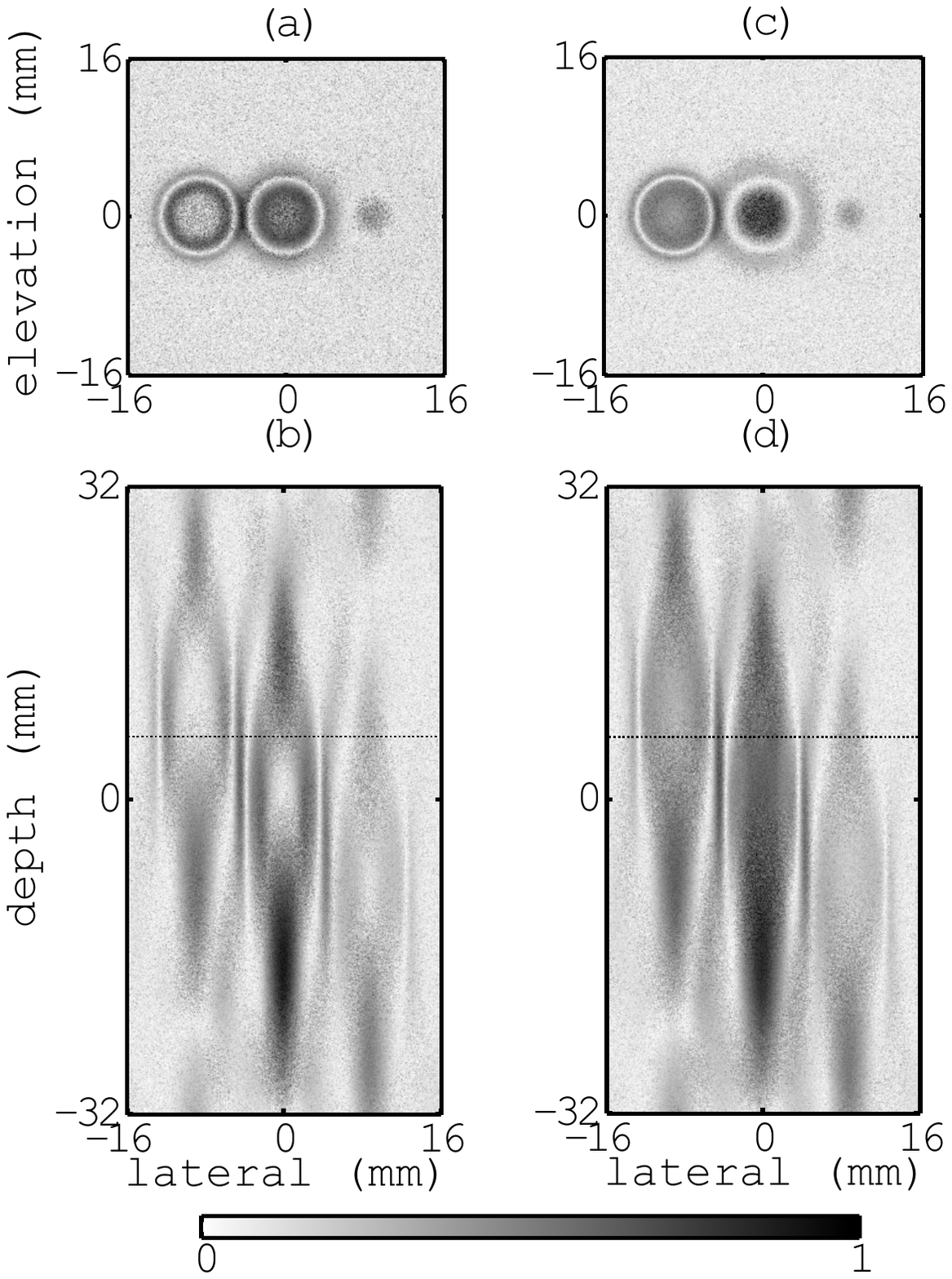}
 \caption{VA images obtained by computational simulations with white noise of
 $\unit[20]{dB}$ SNR. (a) Image
 of phantom 1 seen in the \unit[6]{mm}-transverse plane.
 (b) Phantom 1 seen along depth. (c) Image of phantom 2
 at the \unit[8]{mm}-transverse plane.
 (d) Phantom 2 shown along depth.
The dotted lines in (b) and (d) indicates the position of the images (a) and (b).}
 \label{fig:figure5}
\end{figure}

The VA computational images are generated by convolving the system PSF with the function
which represents the phantom, as given in Eq.~\eqref{eq1}.
A Gaussian white noise of $\unit[20]{dB}$ signal-to-noise ratio (SNR) was added
to the generated images.
The simulated images are presented in Fig.~\ref{fig:figure5}.
It is not possible to recognize any inclusion from the axial image information.

\begin{figure*}[htb!]
\centering
 \includegraphics[width=0.56\linewidth,angle=-90]{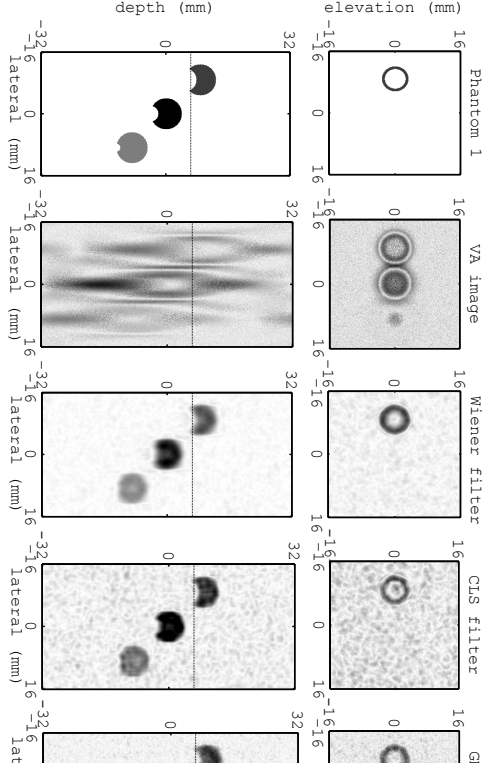}
 \caption{Image restoration results for phantom 2 on both
 the $\unit[6]{mm}$-transverse plane and in-depth.
The filter parameters are $\gamma= 0.2$ (CLS), and $\alpha=0.5$ and $\gamma=1$ (geometric mean). 
 The CLS filter yielded the best boarder preservation,
 while the Wiener filter yielded the most blurred image.
The dotted line indicates where the respective transverse image is placed in depth.}
 \label{fig:figure6}
\end{figure*}

\begin{figure*}[htb!]
\centering
 \includegraphics[width=0.59\linewidth,angle=-90]{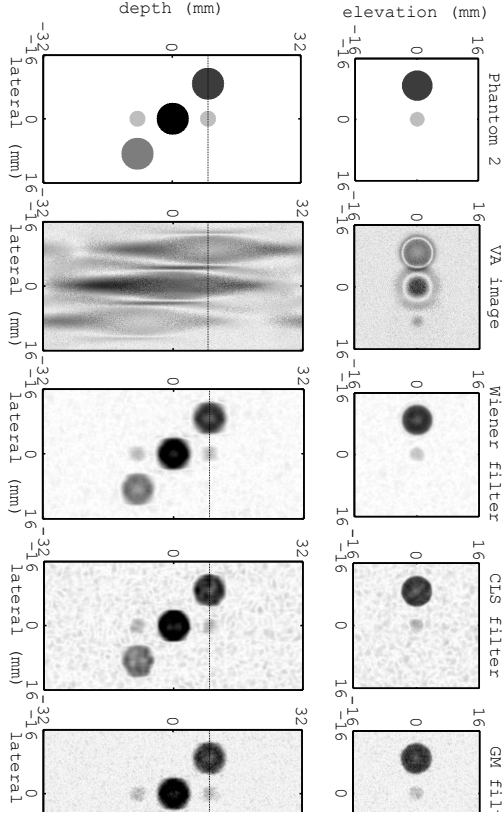}
 \caption{Image restoration results for phantom 2 on both
 the $\unit[8]{mm}$-transverse plane and in-depth.
The filter parameters are $\gamma= 0.2$ (CLS), and $\alpha=0.5$ and $\gamma=1$ (geometric mean). 
 The image restored with the Wiener filter is more blurred than the other results. 
The CLS filter yielded the best edge preservation among the other filters. }
 \label{fig:figure7}
\end{figure*}

The simulated images are restored with the CLS, geometric mean and Wiener filters.
The obtained results for both phantoms are shown in Figs.~\ref{fig:figure6} and~\ref{fig:figure7}.
It is clear that all details in the axial images are recovered by the filters analyzed here.

\begin{figure*}[hbt!]
 \centering
 \subfigure[Phantom 2]{\includegraphics[width=0.28\linewidth,angle=-90]{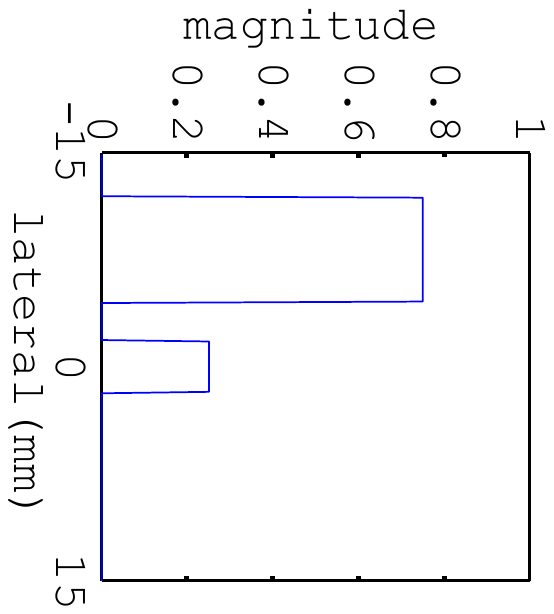}\label{fig:figure8a}}
 \subfigure[VA image]{\includegraphics[width=0.28\linewidth,angle=-90]{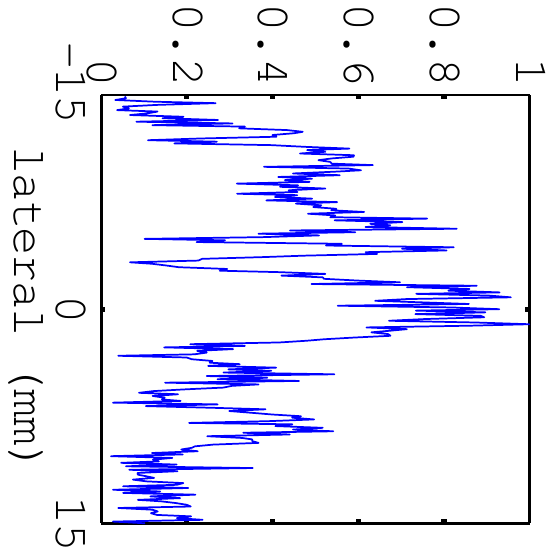}\label{fig:figure8b}}\\
\subfigure[Wiener filter]{\includegraphics[width=.28\linewidth,angle=-90]{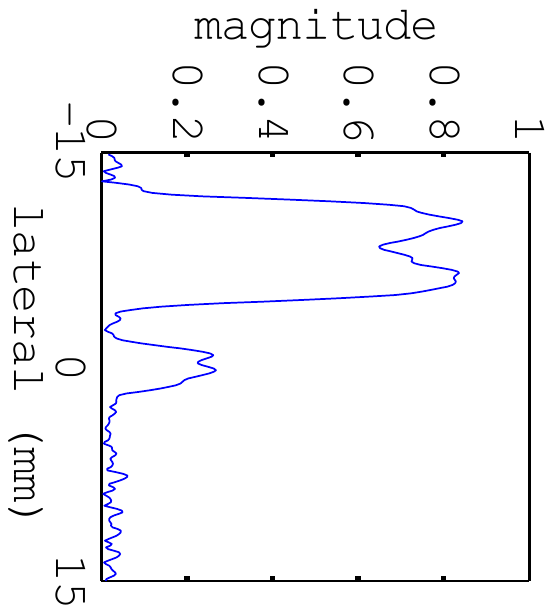}\label{fig:figure8c}}
\subfigure[CLS filter]{\includegraphics[width=.28\linewidth,angle=-90]{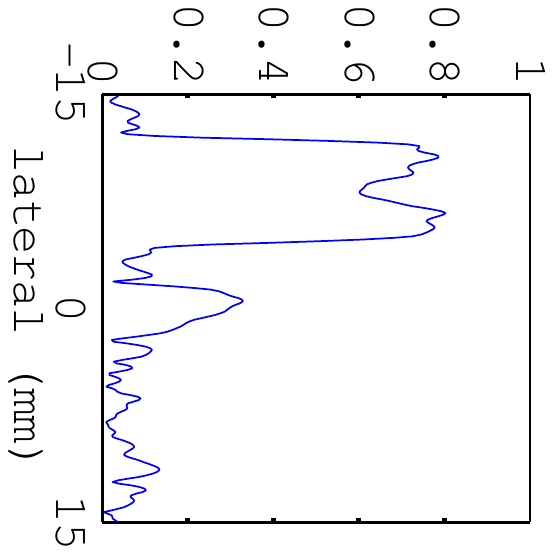}\label{fig:figure8d}}
\subfigure[GM filter]{\includegraphics[width=.28\linewidth,angle=-90]{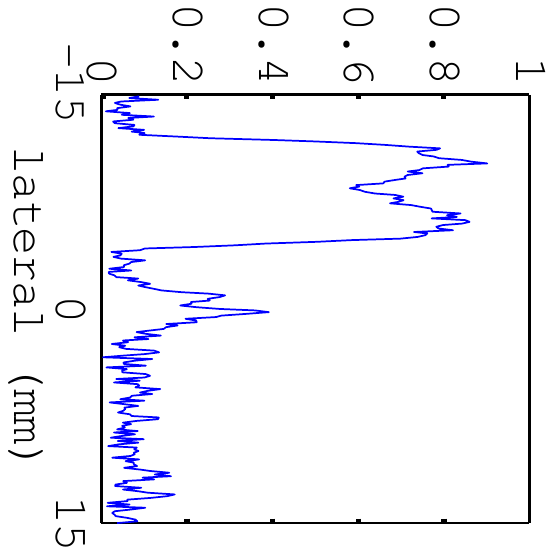}\label{fig:figure8e}}
 \caption{Gray levels obtained for a transverse line at $\unit[8]{mm}$ in depth of phantom 2.
 \subref{fig:figure8a} Transverse line for phantom 2.
 \subref{fig:figure8b} Transverse line for the simulated VA image of phantom 2.
 Transverse lines for the restored images using~\subref{fig:figure8c} Wiener filter, 
 \subref{fig:figure8d} CLS filter and~\subref{fig:figure8e} GM filter.
}
 \label{fig:figure8}
\end{figure*}

In Fig.~\ref{fig:figure8}, the profile of phantom 2  and the corresponding restored images at $\unit[$8$]{mm}$ in depth are displayed in normalized scale. 
The VA image profile in Fig. 7(b) shows a noisy mixture
of two gray levels present in phantom 2.
After the restoration process, the gray levels resemble 
that of phantom 2.

The quantitative analysis is performed using the measures defined in Sec.~\ref{subsec:parameters}.
Deconvolution restoration results for phantoms 1 and 2 are presented in Table~\ref{tab:table1}.
The values indicate good results for the three restoration algorithms
because the ISNR gives \unit[60]{dB} improved with
respect to the raw image.
Moreover, the MSE is lower than $0.02$ and the UIQI is higher than $0.77$ (when this index values one, it indicates 
``perfect'' restoration).
The best restoration result was achieved by the Wiener filter for both phantoms.
We observe that the resulting images are smoother than those obtained 
with the two other filters.
In other words, the restored Wiener image is more blurred and
the object borders are mixed with the background.
Furthermore,  the constrained least-squares filtered images display the best border preservation because 
it is based on the Laplacian operator, which enhances the object borders.
\begin{table*}[ht!]
\caption{
Quantitative results based on the ISNR (improvement in signal-to-noise ratio), 
MSE (mean squared error), and UIQI (universal image quality index) for 
phantoms 1 and 2. 
}
\label{tab:table1}
\begin{center}
\begin{tabular}{l|c|c|c||l|c|c}
\toprule
& \multicolumn{3}{c||}{\small{Phantom 1}} & \multicolumn{3}{c}{\small{Phantom 2}}\\\midrule
\small{Filters} & \small{ISNR} & \small{MSE} & \small{UIQI} & \small{ISNR} & \small{MSE} & \small{UIQI}\\ 
\midrule
\small{Constrained least-squares} & \small{$60.13$} & \small{$0.0192$} & \small{$0.78$} & \small{$59.91$} & \small{$0.0188$} & \small{$0.79$}\\
\small{Geometric mean} & \small{$60.74$}  & \small{$0.0166$} & \small{$0.80$} & \small{$59.83$} &  \small{$0.0191$} &  \small{$0.77$}\\
\small{Wiener} & \small{$61.83$} & \small{$0.0129$} & \small{$0.91$} & \small{$61.03$} & \small{$0.0145$} & \small{$0.88$}\\
\bottomrule
\end{tabular}
\end{center}
\end{table*}
\normalfont

\subsection{Experimental images}
\begin{figure}[hbt!]
\centering
 \includegraphics[width=.6\linewidth,angle=-90]{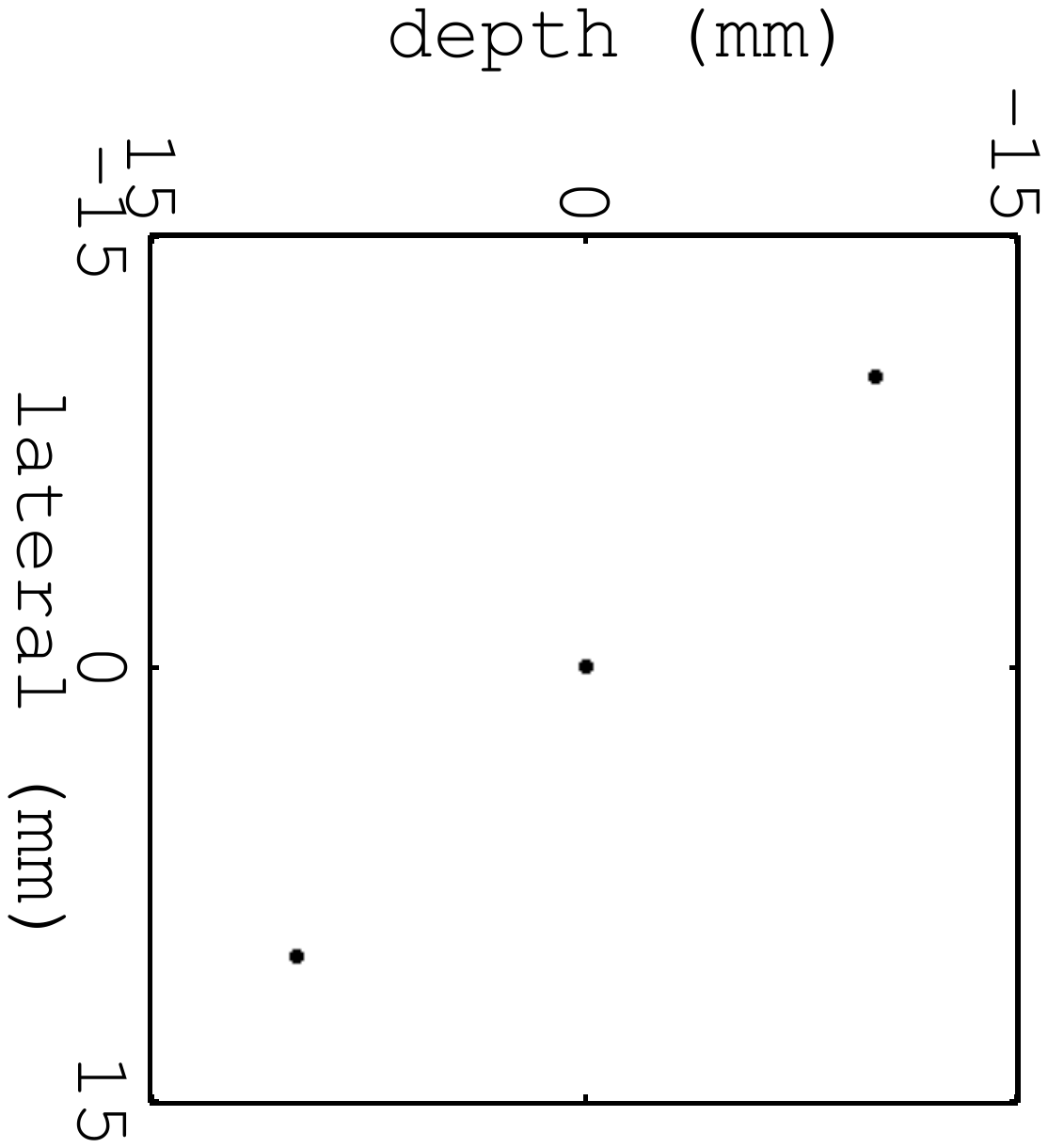}
 \caption{Image of the computational representation of the wire phantom.}
 \label{fig:figure9}
\end{figure}

\begin{figure}[hbt!]
 \centering
 \subfigure[Side view.]{\includegraphics[width=0.42\linewidth]{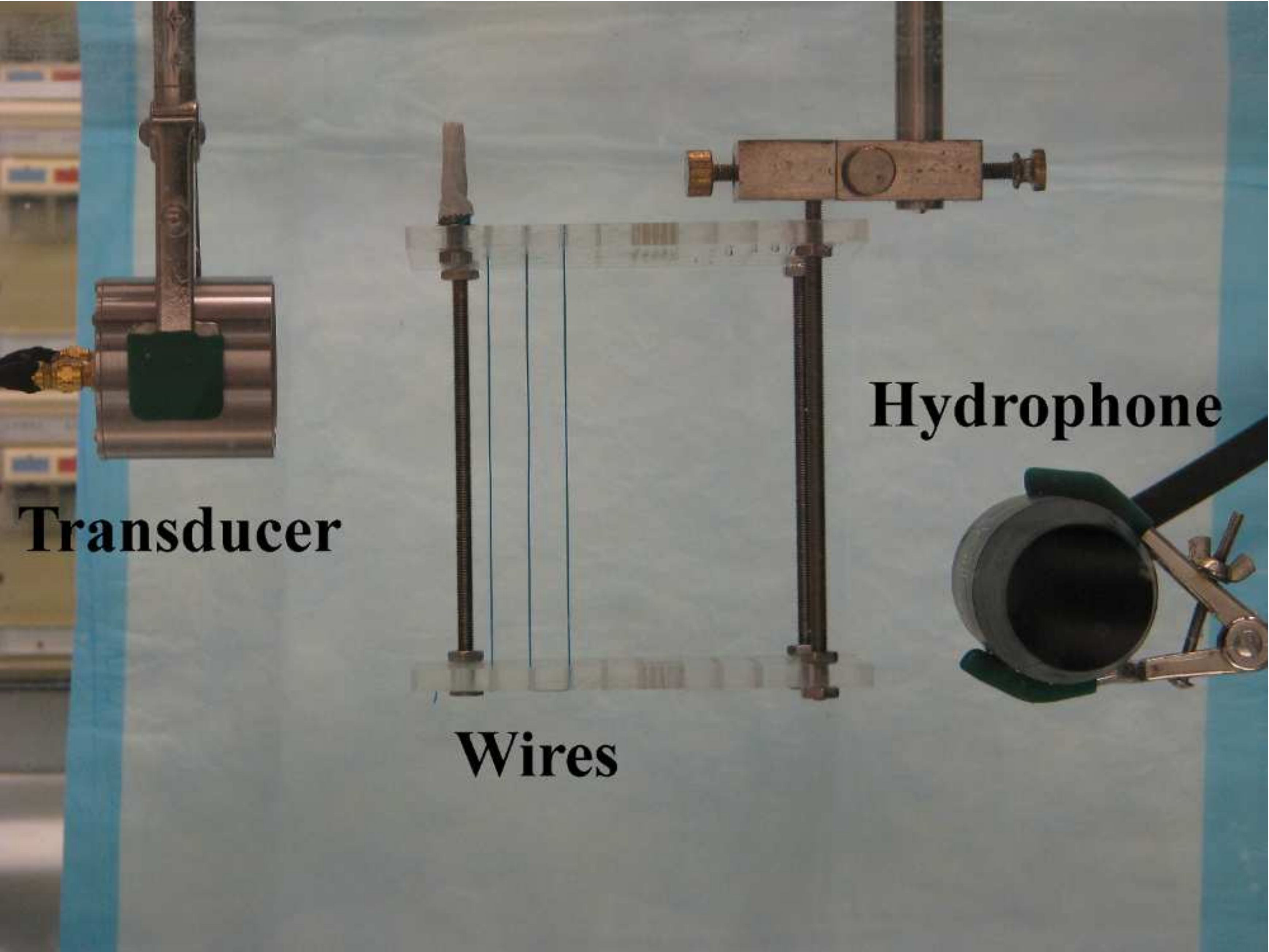}\label{fig:figure10}}\qquad
 \subfigure[Top view.]{\includegraphics[width=0.42\linewidth]{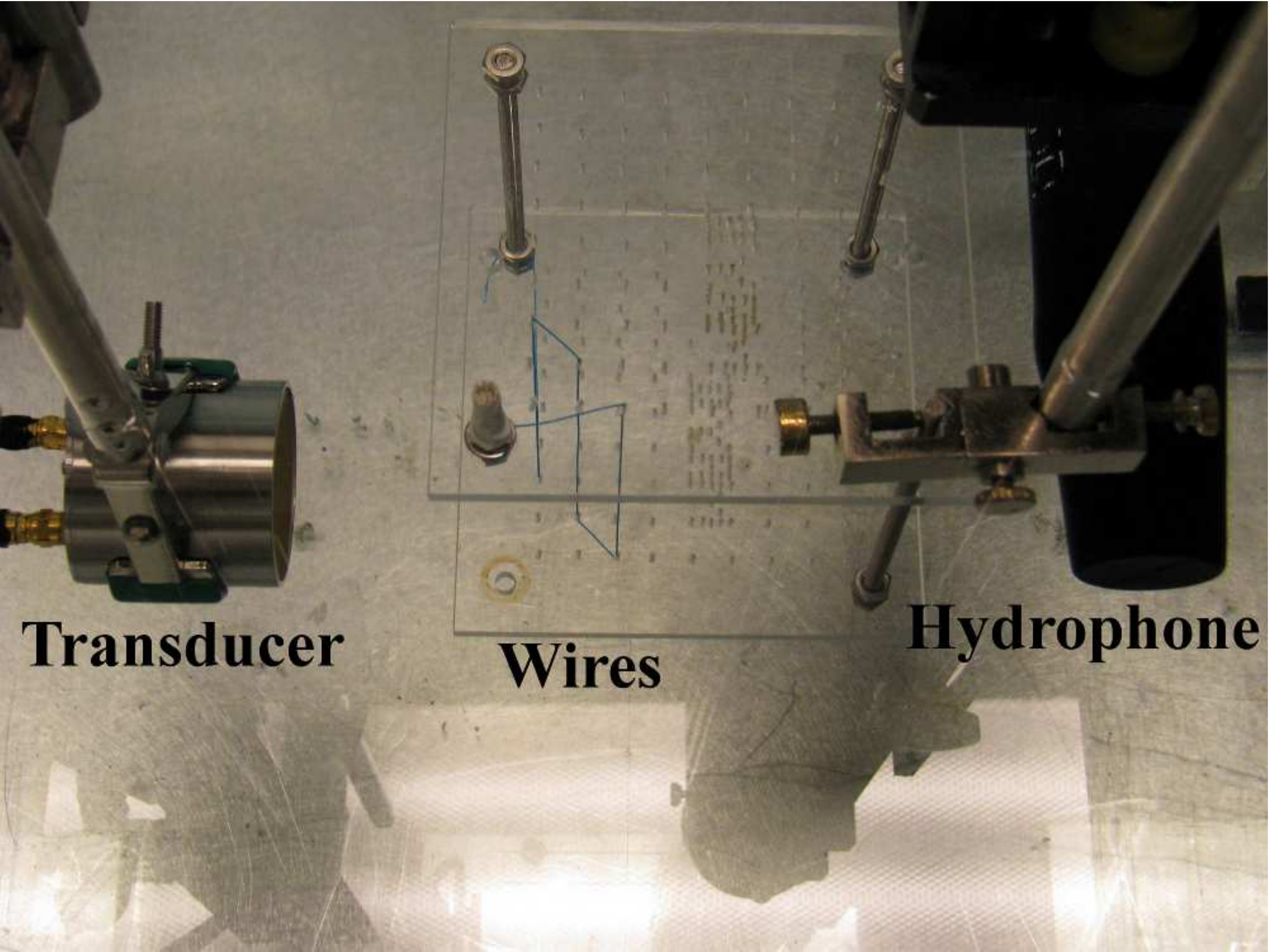}\label{fig:figure11}}
 \subfigure[View from transducer.]{\includegraphics[width=0.42\linewidth]{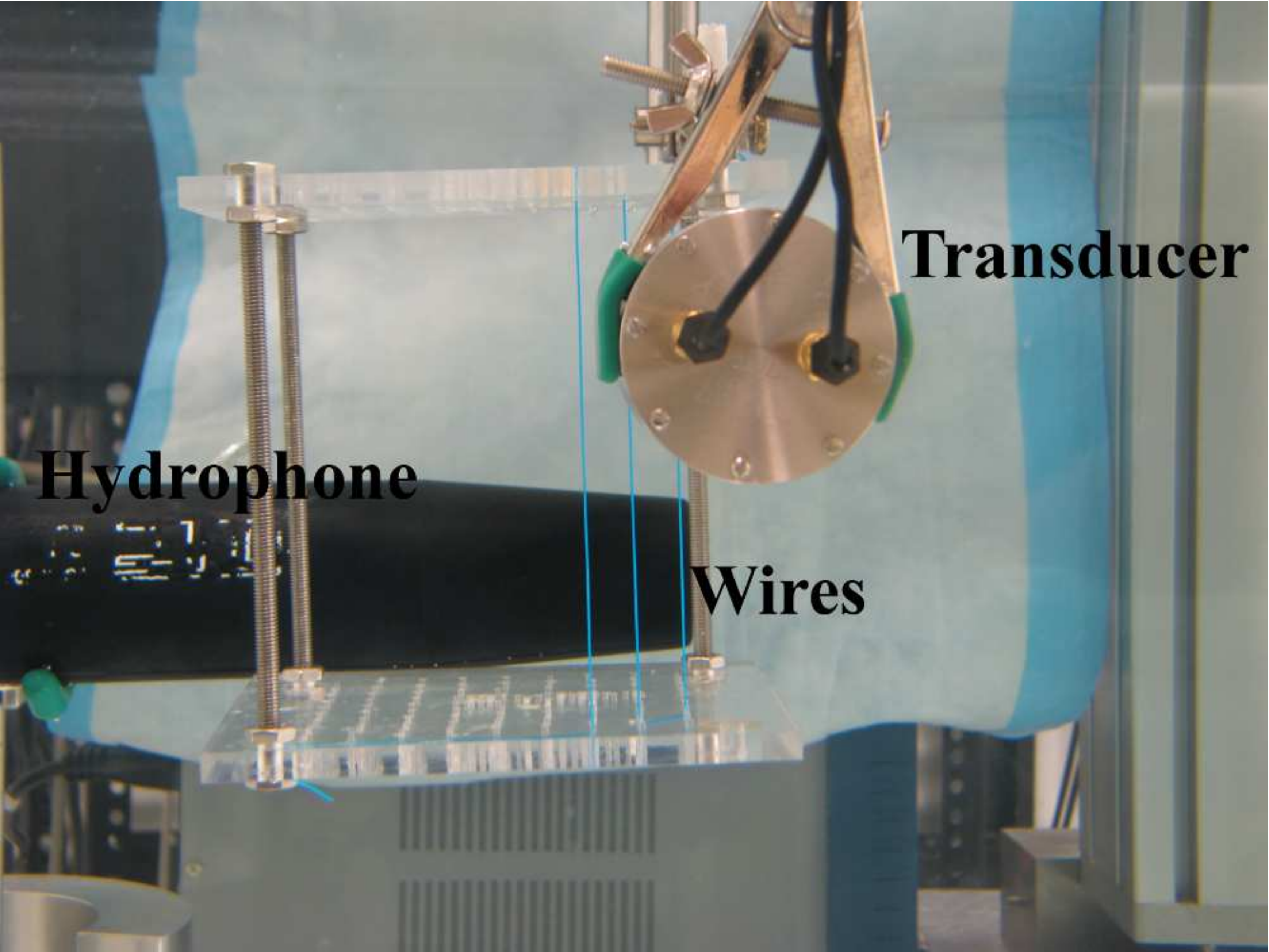}\label{fig:figure12}}
 \caption{VA experimental setup composed of a two-element confocal transducer, a phantom with three stretched thin wires, and a hydrophone. All components are submerged in a water tank.}
 \label{fig:exp_setup}
\end{figure}

To investigate how actual VA images can be restored, an experiment using a VA system is performed and 
the results are presented here.
The VA system consists of the two-element confocal transducer described in
Sec.~\ref{sec:va}, which scans a wire phantom (see Fig.~\ref{fig:figure9}).
The driving frequencies of the transducer elements
are $3.075$ and $\unit[3.125]{MHz}$
and the corresponding difference-frequency is
$\unit[50]{kHz}$.
In turn,
the wire phantom is composed of three stretched \unit[$0.5$]{mm}-diameter wires distributed along the
depth-of-field.
The wires are diagonally arranged and spaced by \unit[$1.4$]{cm}.
The scanned area is \unit[$30 \times 70$]{ mm} and each pixel was taken in steps of \unit[$0.1$]{mm}.
This area encloses all depth-of-field information of the acquired images.
A hydrophone (ITC-6050C, ITC, Santa Barbara, CA, USA) with sensitivity $\unit[-157 \pm 3]{dB~re~1 V/\mu Pa}$ is 
used to detect the generated difference-frequency pressure.
The hydrophone is placed at approximately $\unit[r=20]{cm}$ way from the
farthest wire to the transducer. 
This corresponds to $kr=41$, which can be regarded as the hydrophone
farfield region.
The active element of the hydrophone is a cylinder located in the middle of its conical tip.
This cylinder has a height of $\unit[50]{mm}$ and diameter of $\unit[23]{mm}$.
Its axis is aligned with the the hydrophone long axis.
The hydrophone cannot be considered as a point detector for difference-frequency wavelength of 
few centimeters ($\unit[3.0]{cm}$ at $\unit[50]{kHz}$ in water).
The finite-size effect of the active element (in a linear model) yields an output voltage proportional 
to the spatial average of the detected pressure over the element area plus a signal-independent noise~\cite{boutkedjirt:745}.
Consequently, the hydrophone may add an extra information to the phase of the acquired signal.
Fig.~\ref{fig:exp_setup} shows the experimental VA setup used.

Two schemes are devised to restore the actual VA images.
Firstly, we estimate the line-spread function (LSF) directly from the
experimental image (central wire) and use it in the deconvolution process.
In order to make the experimental LSF more similar to the theoretical one, we applied a spectral
inversion to the image, i.e., first a low-pass filtering is performed and 
then the filtered image is subtracted from the original image~\cite{smith97}.
Secondly, because the magnitude of both the theoretical and the estimated LSF are similar,
 we compose the LSF by combining the magnitude of 
Eq.~\eqref{psf1} and the phase information directly from the experimental data.
Such arrangement was chosen because we found discrepancies between the 
phase of the estimated and the theoretical LSF
based on the radiation force model.
This model seems not to provide the entire phase information present in VA images.
A more complete image formation description needs to take into account parametric and nonlinear scattering phenomena.

\begin{figure}[hbt!]
 \centering
\subfigure[]{\includegraphics[width=0.57\linewidth,angle=-90]{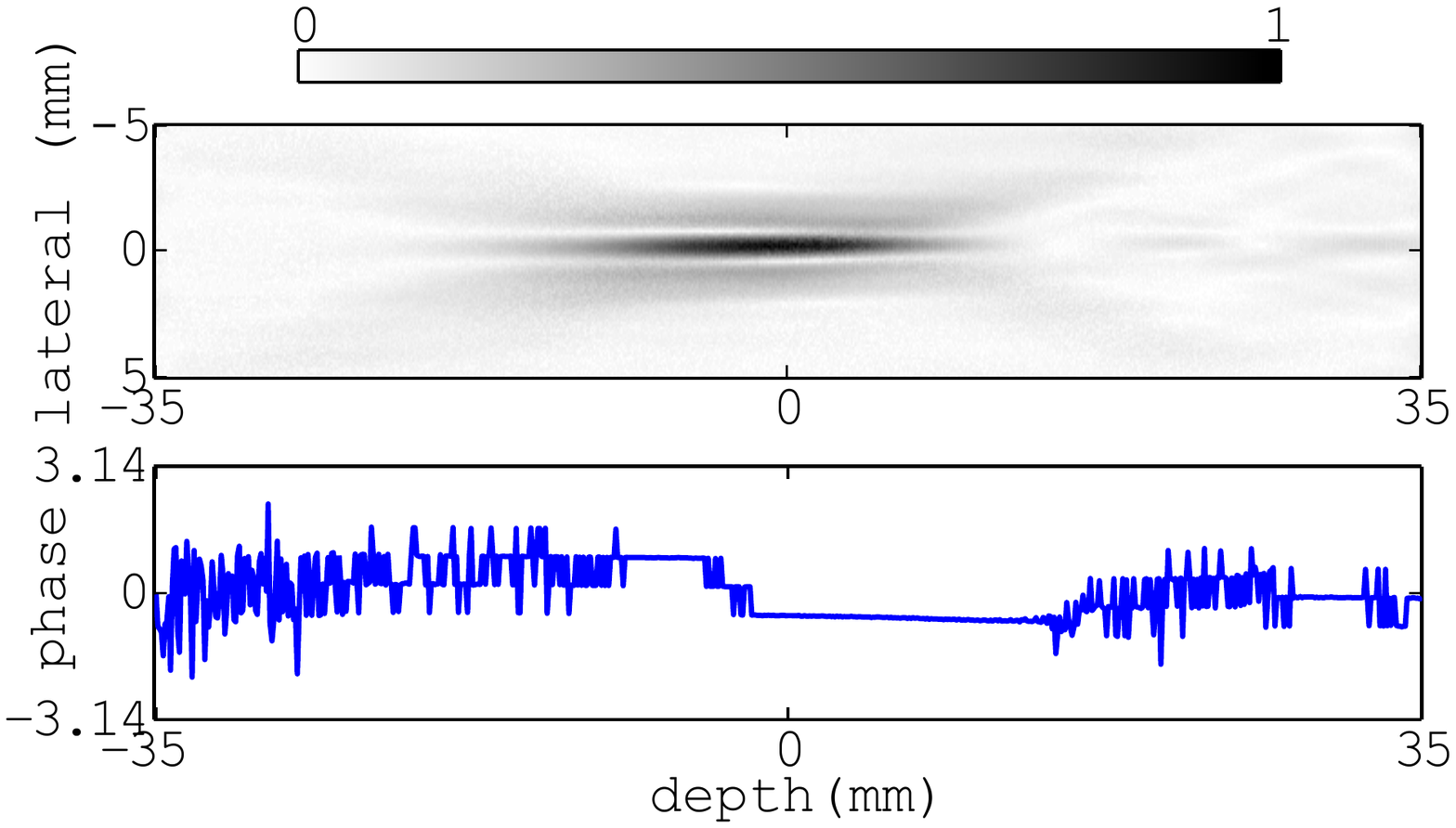}\label{fig:figure13}}
\subfigure[]{\includegraphics[width=0.57\linewidth,angle=-90]{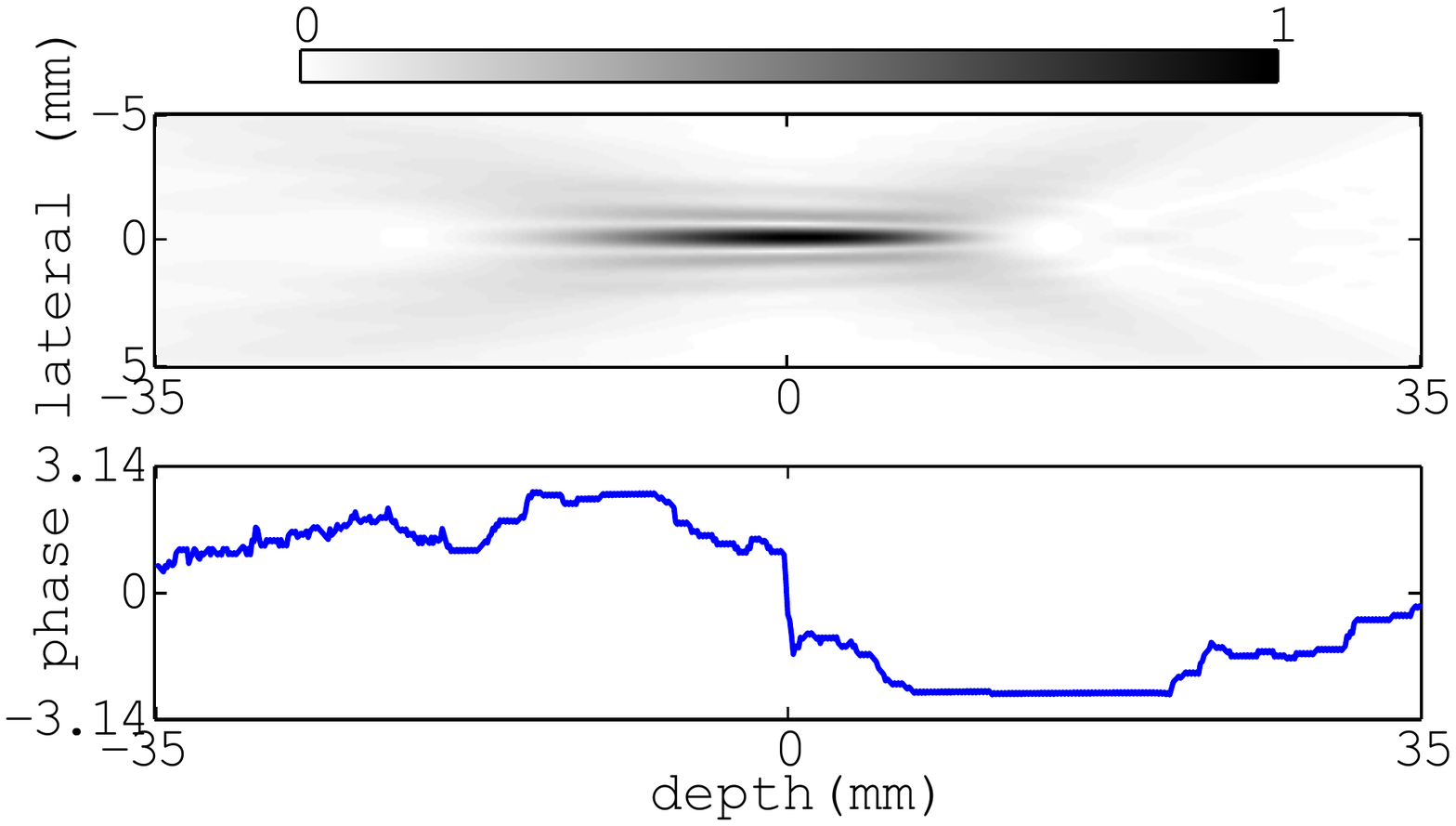}\label{fig:figure14}}
\caption{The magnitudes of (a) the experimental and (b) the theoretical LSFs.
The corresponding phases in radians of the center line along the depth-of-field are also displayed. \label{fig:lsf}}
\end{figure}

The magnitude and phase of the estimated and the theoretical LSFs are exhibited in
Figs.~\ref{fig:figure13} and~\ref{fig:figure14}.
The theoretical LSF was obtained by the convolution of the 3D VA PSF with the 3D wire phantom.
The phase of the experimental LSF is estimated as follows.
Each pixel is formed from an acquired harmonic sequence.
This sequence discretized from the detected signal by the hydrophone at the difference-frequency $\Delta\omega$.
Suppose that the time sequence which represents the $i$th-pixel of the VA image is given by
\begin{equation}
\label{fm}
 x_i[n] = A_i\cos(\Delta \omega n/f_s + \phi_i), \quad n = 0, 1, \dots, N,
\end{equation}
where $A_i>0$ and $\phi_i$ are the is the phase of the signal, $f_s=\unit[1]{MHz}$ is the
sampling frequency, and $N=512$ the number of sampling points. 
By taking the Hilbert transform~\cite{wolfram:hilbert} of $x_i[n]$, one obtains
$\tilde{x}_i[n]  = A_i\sin(\Delta \omega n/f_s + \phi_i).
$
Thus, the representation of the acquired signal in complex notation is
\begin{equation}
x_i[n]+ i\tilde{x}_i[n] = A_i e^{i(\Delta \omega n/f_s + \phi_i)}.
\end{equation}
Therefore, the phase of a sequence which represents the $j$th-pixel with
respect to that of the $i$th-pixel
can be  computed from
\begin{equation}
\phi_j = \phi_i + \frac{i}{A_i A_j N} \sum_{n=0}^N \ln\left[(x_i[n] + i\tilde{x}_i[n])(x_j[n] - i\tilde{x}_j[n])\right].
\end{equation}
All the pixel phases can be obtained with respect to a reference pixel.
The amplitudes $A_i$ and $A_j$ can be calculated by computing
the root mean square of the respective acquired signal.

It can be noticed from Figs.~\ref{fig:figure13} and~\ref{fig:figure14} that 
the theoretical and the experimental LSFs have differences in their phase contents.
The lack of information in the LSF phase leads to unsatisfactory restoration results with several artifacts.
The missing information in the phase is mostly related to the fact that the
image formation model based on Eq.~(\ref{psf1}) does not consider the nonlinear interaction of the primary beams.
Moreover, the finite-size effect of the hydrophone active element was not considered in the theoretical model.

\begin{figure}[hbt!]
 \centering
 \subfigure[]{\includegraphics[width=\linewidth]{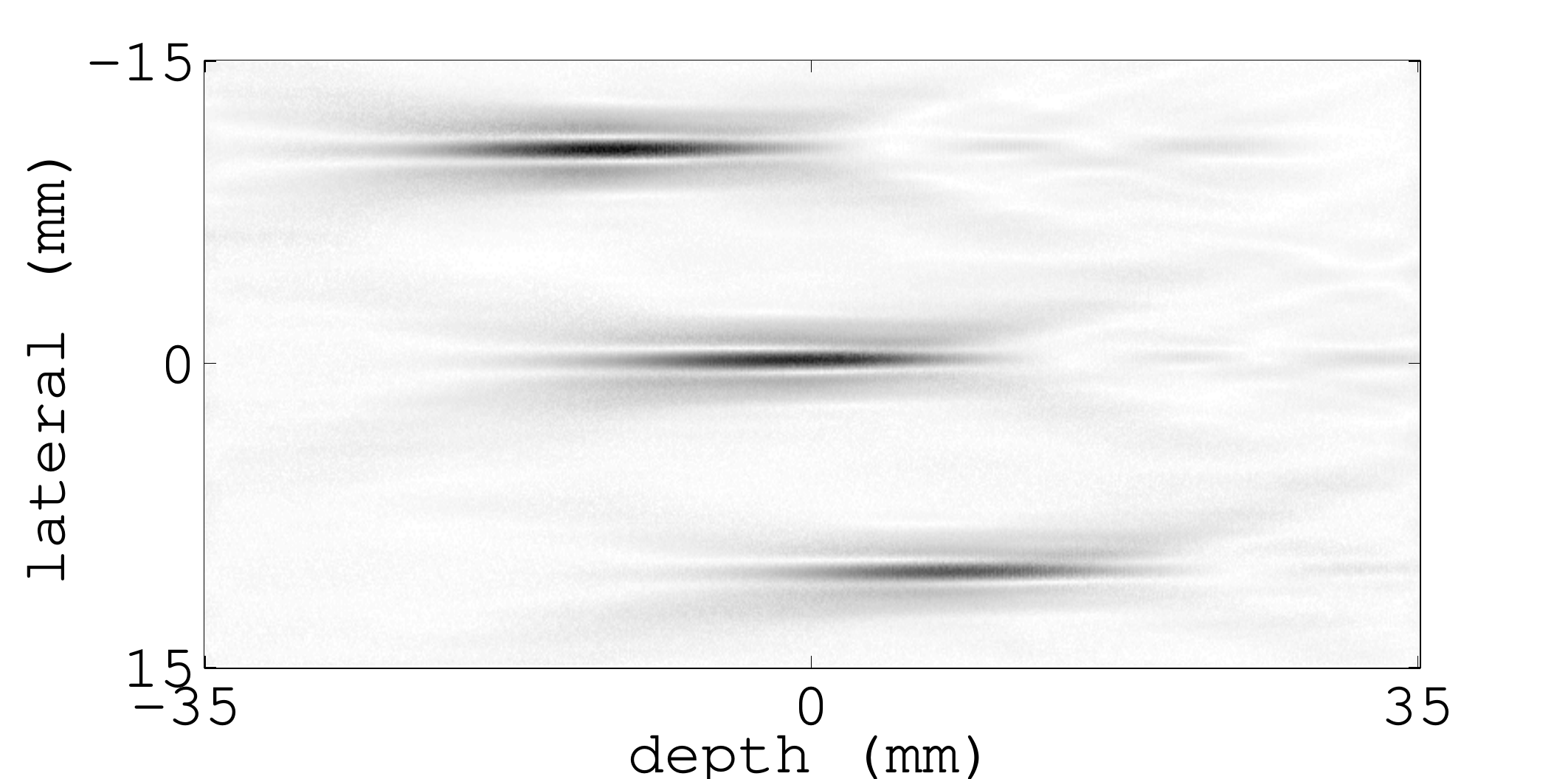}\label{fig:figure15}}
 \subfigure[]{\includegraphics[width=\linewidth]{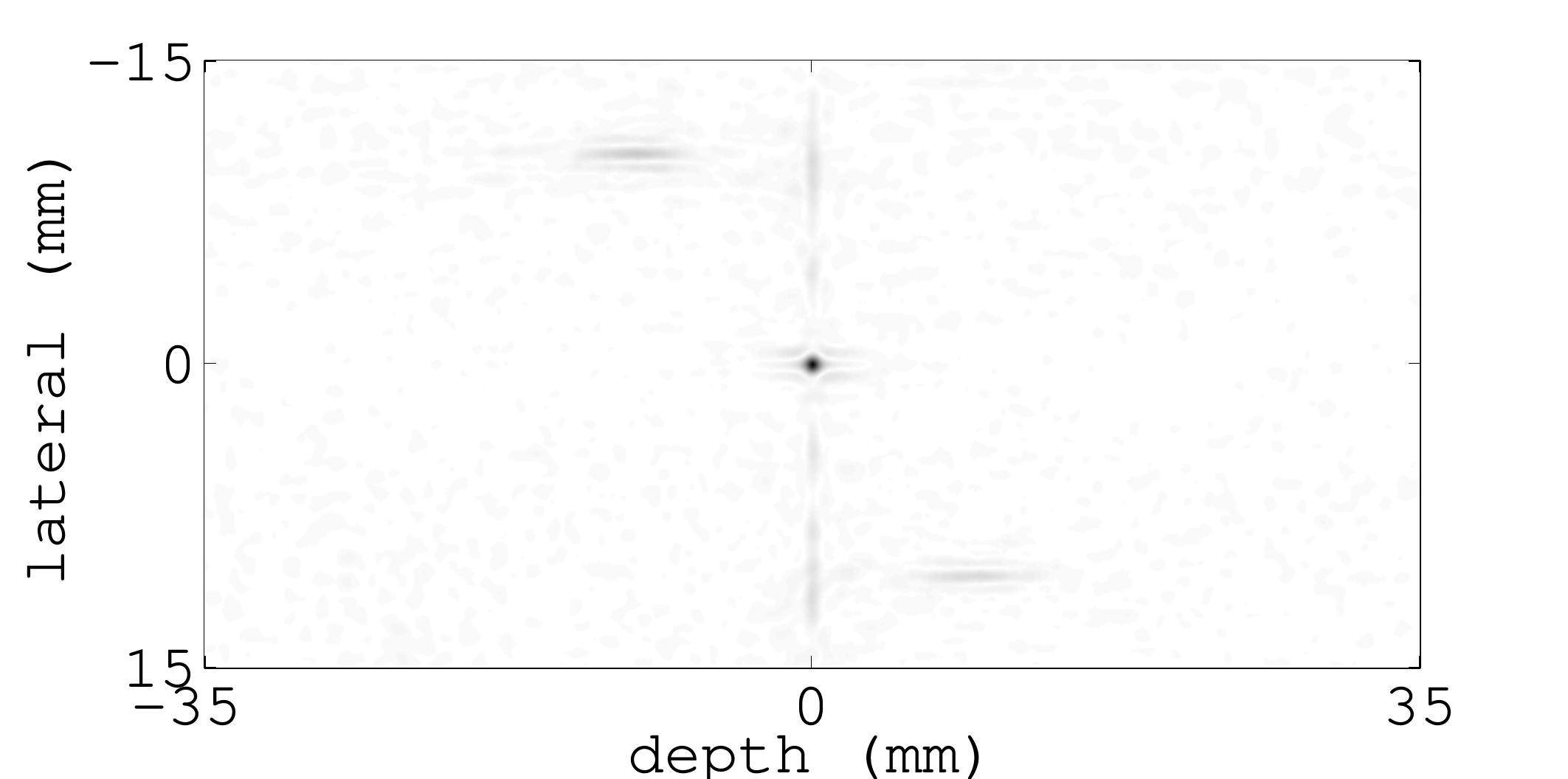}\label{fig:figure16}}
 \subfigure[]{\includegraphics[width=\linewidth]{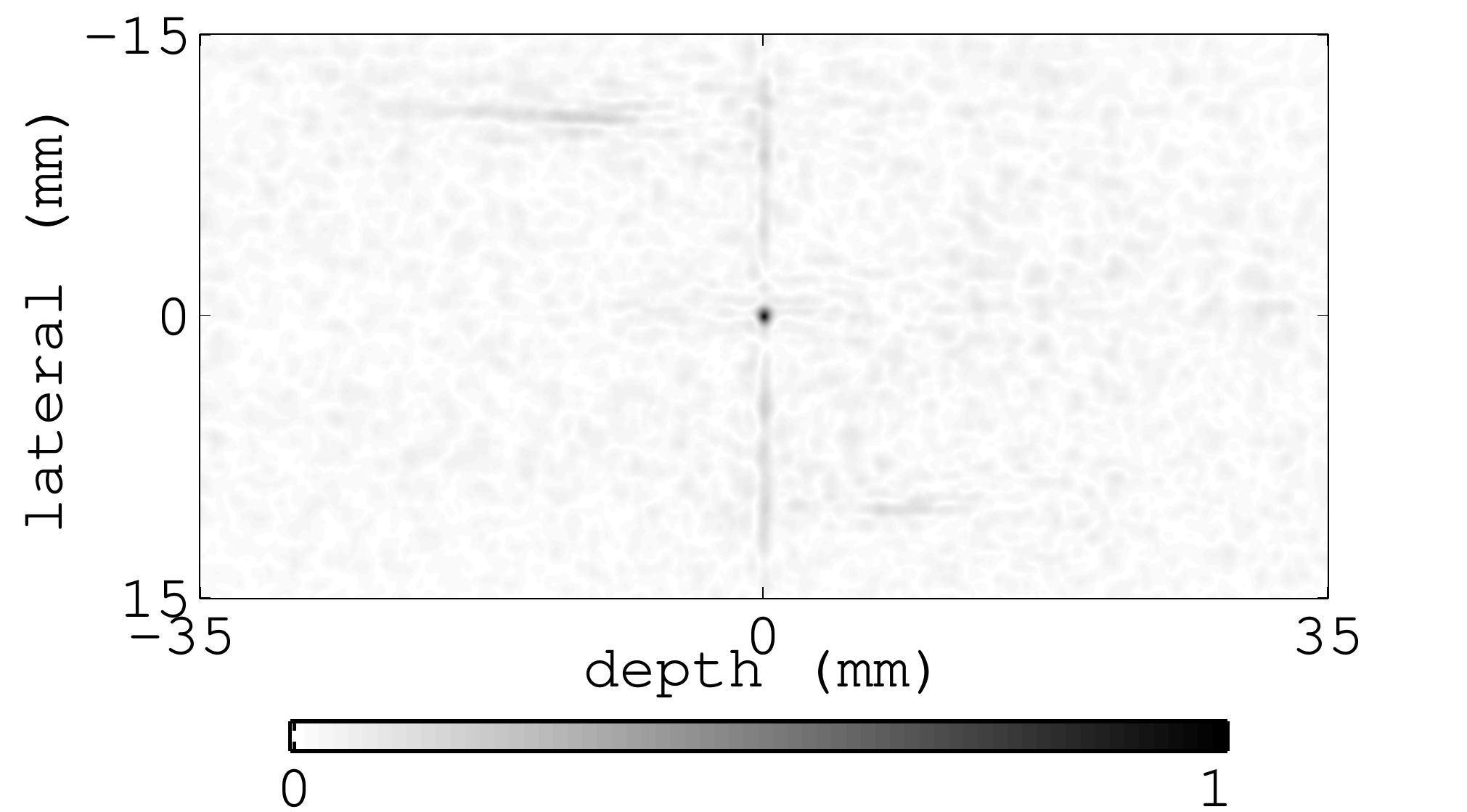}\label{fig:figure17}}
  \subfigure[]{\includegraphics[width=\linewidth]{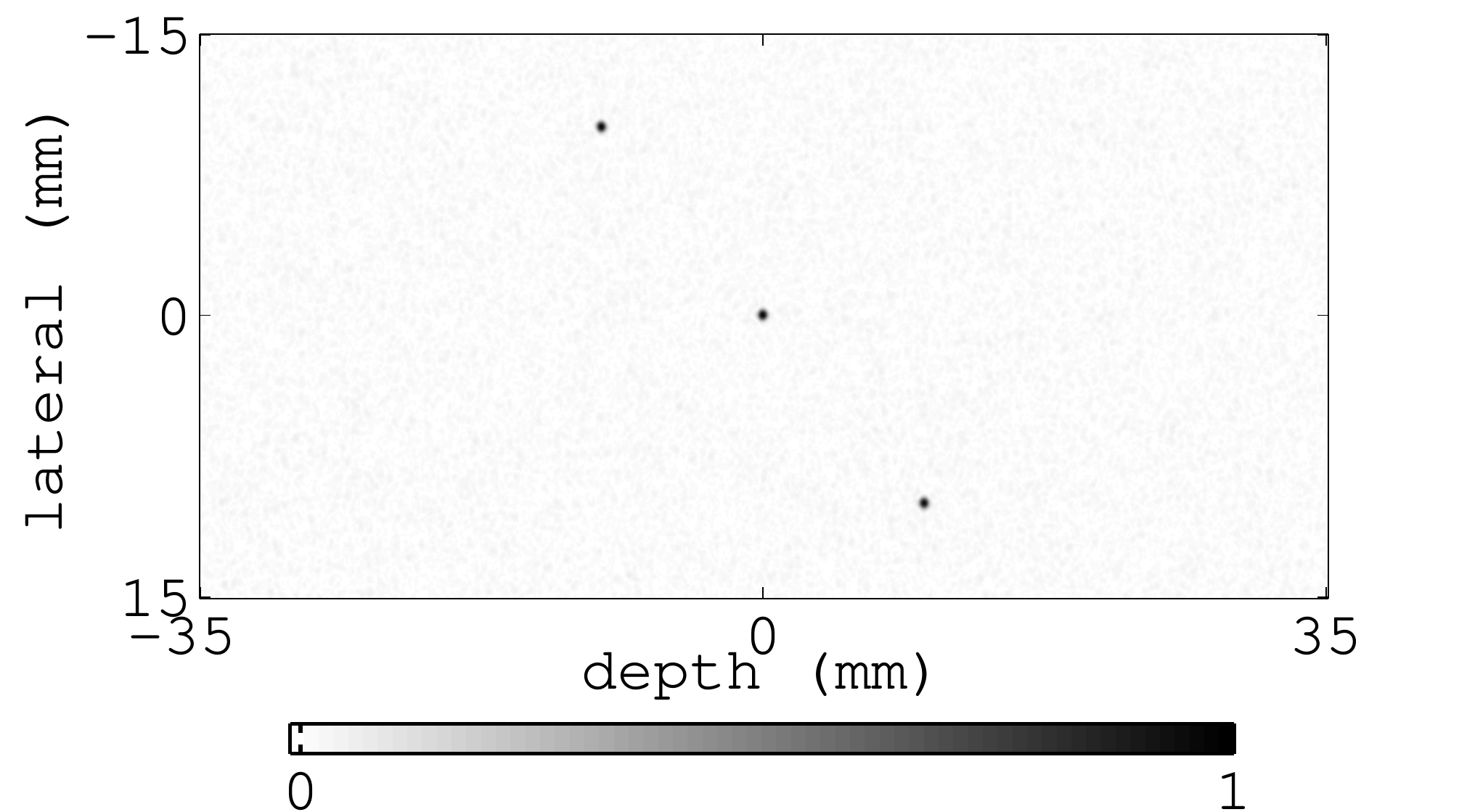}\label{fig:figure18}}
 \caption{Image restoration using the CLS filter with $\gamma=0.01$. The raw VA image is shown in (a).
The restoration results are presented for the experimental (b) and the composed (c) LSFs.
 The deconvolution for the computer generated VA image is shown in (d). }
 \label{fig:image}
\end{figure}

The result obtained by estimating the LSF 
from the experimental image, is presented in Fig.~\ref{fig:figure16}, while
the result using the composed LSF is shown in Fig.~\ref{fig:figure17}.
Both methods yielded similar restored images.
Despite presenting two blurred wires (top left and bottom right), the restored images show aspects of the 
imaged wires, including location, geometric shape and contrast.
Some aspects may have contributed for the occurrence of artifacts in the restored images.
The deconvolution filters used here usually produce undesirable artifacts during the 
process of deconvolution because of the ill-posed nature of the image restoration problem, i.e.,
it may have more than one solution and the solutions depend discontinuously on the initial data.
Therefore, perturbations in the VA image due to the noise, may cause undesired variations in the restoration results.
This can be noticed by comparing the result of a complete simulated deconvolution
process presented in Fig.~\ref{fig:figure18}, where no artifacts are produced, because the
whole process is controlled, from the image formation up until its restoration.
That is, the same LSF is used for both VA image formation and restoration.
Furthermore, it is important to note that an extra information may appear in the phase of the detected signal
caused by the acoustical path difference from the wire positions to the hydrophone.

\section{Summary and conclusions}
\label{sec:conclusions}
In this paper, we have presented the results of three restoration techniques (Wie\-ner,
CLS and geometric mean filters) applied in the restoration of VA images.
The restoration algorithms based on the theoretical PSF were applied to computationally simulated images, in which the visual information
of the objects was lost.
The performance of the algorithms were assessed both quantitatively and visually.
The deconvolution of simulated images was capable of restoring the size, shape, position and contrast of the objects present in the images for all filters used.

The CLS filter was employed in restoring an actual VA image of three stretched wires.
In this case, two LSFs were used, one was estimated from the image, while the other 
combines the phase of the estimated LSF with the magnitude of the theoretical one.
The wire positions and partial information about their shape and contrast can be recognized
in the images restored with both experimental and combined LSFs.

The restoration of the experimental VA image with the theoretical LSF based on 
Eq.~\eqref{psf1} yielded a poor visual restoration retrieve and
thus was not presented here.
Yet the magnitudes of both theoretical and experimental LSFs matched, 
the phases showed distinctly different contents.
This hints that the adopted theoretical model for the PSF lacks the appropriate phase information of VA images.
Additional information may come from the difference-frequency generation based on 
the nonlinear interaction of the primary waves~\cite{silva:1326,silva:2011,westervelt:535}
and possibly also on the finite-size effect of the hydrophone active element.

This study represents a preliminary step in understanding how application of deconvolution filters can be used to 
restore the resolution and contrast for VA images. The optimal restoration filter may depend on the object 
being imaged with to improve resolution and contrast. 
Future studies would be necessary to understand and 
validate how different filters and modeling of the PSF would affect the restoration of VA images of simple 
phantoms and soft tissues.

\section{Acknowledgments}
This work was partially supported by FAPESP and Grants 306697/2010-6, 
477653/2010-3 CNPq (Brazilian agencies).


\section*{References}

\small

\begin{thebibliography}{25}
\expandafter\ifx\csname natexlab\endcsname\relax\def\natexlab#1{#1}\fi
\providecommand{\bibinfo}[2]{#2}
\ifx\xfnm\relax \def\xfnm[#1]{\unskip,\space#1}\fi
\bibitem[{Fatemi and Greenleaf(1998)}]{fatemi98}
\bibinfo{author}{M.~Fatemi}, \bibinfo{author}{J.~F. Greenleaf},
\newblock \bibinfo{title}{Ultrasound-stimulated vibro-acoustic
  spectrography},
\newblock \bibinfo{journal}{Science} \bibinfo{volume}{280}
  (\bibinfo{year}{1998}) \bibinfo{pages}{82--85}.
\bibitem[{Greenleaf et~al.(1998)Greenleaf, Ehman, Fatemi, and
  Muthupillai}]{greenleaf98}
\bibinfo{author}{J.~F. Greenleaf}, \bibinfo{author}{R.~L. Ehman},
  \bibinfo{author}{M.~Fatemi}, \bibinfo{author}{R.~Muthupillai},
\newblock \bibinfo{title}{Imaging {E}lastic {P}roperties of {T}issue},
\newblock in: \bibinfo{booktitle}{Ultrasound in {M}edicine}, Medical {S}ciences
  {S}eries, \bibinfo{publisher}{Taylor \& Francis}, \bibinfo{year}{1998}, pp.
  \bibinfo{pages}{263--277}.
\bibitem[{Fatemi and Greenleaf(2000)}]{fatemi00}
\bibinfo{author}{M.~Fatemi}, \bibinfo{author}{J.~F. Greenleaf},
\newblock \bibinfo{title}{Probing the dynamics of tissue at low frequencies
  with the radiation force of ultrasound},
\newblock \bibinfo{journal}{{P}hys. {M}ed. {B}iol.} \bibinfo{volume}{45}
  (\bibinfo{year}{2000}) \bibinfo{pages}{14491464}.
\bibitem[{Fatemi et~al.(2003)Fatemi, Manduca, and Greenleaf}]{fatemi03}
\bibinfo{author}{M.~Fatemi}, \bibinfo{author}{A.~Manduca},
  \bibinfo{author}{J.~F. Greenleaf},
\newblock \bibinfo{title}{Imaging elastic properties of biological tissues by
  low-frequency harmonic vibration},
\newblock \bibinfo{journal}{{P}roc. {IEEE}} \bibinfo{volume}{91}
  (\bibinfo{year}{2003}) \bibinfo{pages}{1503--1517}.
\bibitem[{Fatemi et~al.(2002)Fatemi, Wold, Alizad, and Greenleaf}]{fatemi01}
\bibinfo{author}{M.~Fatemi}, \bibinfo{author}{L.~E. Wold},
  \bibinfo{author}{A.~Alizad}, \bibinfo{author}{J.~F. Greenleaf},
\newblock \bibinfo{title}{Vibro-acoustic tissue mamography},
\newblock \bibinfo{journal}{{IEEE} {T}rans. {M}ed. {I}maging}
  \bibinfo{volume}{21} (\bibinfo{year}{2002}) \bibinfo{pages}{1--8}.
\bibitem[{Alizad et~al.(2004{\natexlab{a}})Alizad, Fatemi, Wold, and
  Greenleaf}]{wold04}
\bibinfo{author}{A.~Alizad}, \bibinfo{author}{M.~Fatemi},
  \bibinfo{author}{L.~E. Wold}, \bibinfo{author}{J.~F. Greenleaf},
\newblock \bibinfo{title}{Performance of vibro-acoustography in detecting
  microcalcifications in excised human breast tissue: a study of 74 tissues
  samples},
\newblock \bibinfo{journal}{{IEEE} {T}rans. {M}ed. {I}maging}
  \bibinfo{volume}{23} (\bibinfo{year}{2004}{\natexlab{a}})
  \bibinfo{pages}{307--312}.
\bibitem[{Alizad et~al.(2004{\natexlab{b}})Alizad, Wold, Greenleaf, and
  Fatemi}]{alizad04}
\bibinfo{author}{A.~Alizad}, \bibinfo{author}{J.~E. Wold},
  \bibinfo{author}{J.~F. Greenleaf}, \bibinfo{author}{M.~Fatemi},
\newblock \bibinfo{title}{Imaging mass lesions by vibro-acoustography:
  modeling and experiments},
\newblock \bibinfo{journal}{{IEEE} {T}rans. {M}ed. {I}maging}
  \bibinfo{volume}{23} (\bibinfo{year}{2004}{\natexlab{b}})
  \bibinfo{pages}{1087--1093}.
\bibitem[{Calle et~al.(2001)Calle, Remenieras, Matar, Defontaine, Gomez, and
  Patat}]{calle01}
\bibinfo{author}{S.~Call\'e}, \bibinfo{author}{J.~P. Remenieras},
  \bibinfo{author}{O.~B. Matar}, \bibinfo{author}{M.~Defontaine},
  \bibinfo{author}{M.~A. Gomez}, \bibinfo{author}{F.~Patat},
\newblock \bibinfo{title}{Application of vibro-acoustography to bone
  elasticity imaging},
\newblock in: \bibinfo{booktitle}{Proc. IEEE Ultrason. Symp.},
  \bibinfo{publisher}{IEEE-UFFC}, \bibinfo{year}{2001}, pp.
  \bibinfo{pages}{1601--1604}.

\bibitem[{Urban et~al.(2011) Urban, Alizad, Aquino, Greenleaf, and
  Fatemi}]{urban:2011}
\bibinfo{author}{M. W.~Urban}, \bibinfo{author}{A.~ Alizad},
  \bibinfo{author}{W. Aquino},\bibinfo{author}{J.~F. Greenleaf}, \bibinfo{author}{M.~Fatemi},
\newblock \bibinfo{title}{A review of vibro-acoustography and its applications in medicine},
\newblock \bibinfo{journal}{Current Med. Imaging Rev.}
  \bibinfo{volume}{7} (\bibinfo{year}{2011})
  \bibinfo{pages}{350-359}.


\bibitem[{Silva et~al.(2008)Silva, Mitri, and Fatemi}]{silva:1326}
\bibinfo{author}{G.~T. Silva}, \bibinfo{author}{F.~G. Mitri},
  \bibinfo{author}{M.~Fatemi},
\newblock \bibinfo{title}{Analysis of the difference-frequency wave generated
  by the interaction of two axisymmetric and co-focused ultrasound beams},
\newblock in: \bibinfo{booktitle}{Proc. IEEE Ultrason. Symp.},
  \bibinfo{publisher}{IEEE-UFFC}, \bibinfo{year}{2008}, pp.
  \bibinfo{pages}{1326--1329}.
\bibitem[{Silva and Mitri(2011)}]{silva:2011}
\bibinfo{author}{G.~T. Silva}, \bibinfo{author}{F.~G. Mitri},
\newblock \bibinfo{title}{Difference-frequency generation in
  vibro-acoustography},
\newblock \bibinfo{journal}{Phys. Med. Biol.}
  \bibinfo{volume}{56} (\bibinfo{year}{2011}) \bibinfo{pages}{5985--5993}.
\bibitem[{Westervelt(1963)}]{westervelt:535}
\bibinfo{author}{P.~J. Westervelt},
\newblock \bibinfo{title}{Parametric acoustic array},
\newblock \bibinfo{journal}{J. Acoust. Soc. Am.} \bibinfo{volume}{35}
  (\bibinfo{year}{1963}) \bibinfo{pages}{535--537}.
\bibitem[{Urban et~al.(2006)Urban, Silva, Fatemi, and Greenleaf}]{urban06}
\bibinfo{author}{M.~W. Urban}, \bibinfo{author}{G.~T. Silva},
  \bibinfo{author}{M.~Fatemi}, \bibinfo{author}{J.~F. Greenleaf},
\newblock \bibinfo{title}{Multifrequency vibro-acoustography},
\newblock \bibinfo{journal}{{IEEE} {T}rans. {M}ed. {I}maging}
  \bibinfo{volume}{25} (\bibinfo{year}{2006}) \bibinfo{pages}{1284--1295}.
\bibitem[{Fatemi and Greenleaf(1999)}]{fatemi99}
\bibinfo{author}{M.~Fatemi}, \bibinfo{author}{J.~F. Greenleaf},
\newblock \bibinfo{title}{Vibro-acoustography: an imaging modality based on
  ultrasound-stimulated acoustic emission},
\newblock \bibinfo{journal}{{P}roc. {N}atl. {A}cad. {S}ci. {USA}}
  \bibinfo{volume}{96} (\bibinfo{year}{1999}) \bibinfo{pages}{6603--6608}.
\bibitem[{Chen et~al.(2004)Chen, Fatemi, Kinnick, and Greenleaf}]{chen04}
\bibinfo{author}{S.~Chen}, \bibinfo{author}{M.~Fatemi},
  \bibinfo{author}{R.~Kinnick}, \bibinfo{author}{J.~F. Greenleaf},
\newblock \bibinfo{title}{Comparison of stress field forming methods
  for vibro-acoustography},
\newblock \bibinfo{journal}{{IEEE} {T}rans. {U}ltrason. {F}erroelectr. {F}req.
  {C}ontrol} \bibinfo{volume}{51} (\bibinfo{year}{2004})
  \bibinfo{pages}{313--321}.
\bibitem[{Silva et~al.(2004)Silva, Greenleaf, and Fatemi}]{glauber04}
\bibinfo{author}{G.~T. Silva}, \bibinfo{author}{J.~F. Greenleaf},
  \bibinfo{author}{M.~Fatemi},
\newblock \bibinfo{title}{Linear arrays for vibro-acoustography: a
  numerical simulation study},
\newblock \bibinfo{journal}{Ultrason. Imaging} \bibinfo{volume}{26}
  (\bibinfo{year}{2004}) \bibinfo{pages}{1--17}.
\bibitem[{Urban et~al.(2011)Urban, Chalek, Kinnick, Kinter, Haider, Greenleaf,
  Thomenius, and Fatemi}]{urban11}
\bibinfo{author}{M.~Urban}, \bibinfo{author}{C.~Chalek},
  \bibinfo{author}{R.~Kinnick}, \bibinfo{author}{T.~Kinter},
  \bibinfo{author}{B.~Haider}, \bibinfo{author}{J.~Greenleaf},
  \bibinfo{author}{K.~Thomenius}, \bibinfo{author}{M.~Fatemi},
\newblock \bibinfo{title}{Implementation of vibro-acoustography on a clinical
  ultrasound system},
\newblock \bibinfo{journal}{IEEE Trans. Ultrason. Ferroelectr. Freq. Control}
  \bibinfo{volume}{58} (\bibinfo{year}{2011}) \bibinfo{pages}{1169--81}.

\bibitem[{Kamimura et~al.(2012) Kamimura, Urban, Carneiro, Fatemi, and Alizad}]{kamimura:}
\bibinfo{author}{H. A. S. Kamimura},
\bibinfo{author}{M.~W.~Urban}, 
\bibinfo{author}{A. A. O. Carneiro},
\bibinfo{author}{M.~Fatemi},
\bibinfo{author}{A. Alizad},
\newblock \bibinfo{title}{Vibro-acoustography beam formation with
reconfigurable arrays},
\newblock \bibinfo{journal}{IEEE Trans. Ultrason. Ferroelectr. Freq. Control} (\bibinfo{year}{2012}), accepted for publication.

\bibitem[{Ding(2000)}]{ding:2759}
\bibinfo{author}{D. Ding},
\newblock \bibinfo{title}{A simplified algorithm for the second-order sound fields},
\newblock \bibinfo{journal}{J. Acoust. Soc. Am.} \bibinfo{volume}{108}
  (\bibinfo{year}{2000}) \bibinfo{pages}{2759--2764}.
\bibitem[{Gonzalez and Woods(2008)}]{gonzalezcap5}
\bibinfo{author}{R.~C. Gonzalez}, \bibinfo{author}{R.~E. Woods},
\newblock \bibinfo{title}{Digital {I}mage {P}rocessing}, \bibinfo{edition}{3} ed., 
  \bibinfo{publisher}{Prentice Hall}, \bibinfo{year}{2008}, ch. 5.
\bibitem[{Hunt(1973)}]{hunt73}
\bibinfo{author}{B.~R. Hunt},
\newblock \bibinfo{title}{The {A}pplication of constrained least squares
  estimation to image restoration by digital computer},
\newblock \bibinfo{journal}{{IEEE} {T}rans. {C}omput.} \bibinfo{volume}{C-22}
  (\bibinfo{year}{1973}) \bibinfo{pages}{805--812}.
\bibitem[{Perciano et~al.(2008)Perciano, Mascarenhas, Frery, and
  Silva}]{perciano08}
\bibinfo{author}{T.~Perciano}, \bibinfo{author}{N.~D.~A. Mascarenhas},
  \bibinfo{author}{A.~C. Frery}, \bibinfo{author}{G.~T. Silva},
\newblock \bibinfo{title}{Restoration of vibro-acoustography images},
\newblock in: \bibinfo{booktitle}{SAC '08: {P}roc. {ACM} {S}ymp. {A}ppl.
  {C}omput.}, pp. \bibinfo{pages}{1762--1763} (\bibinfo{year}{2008}).
\bibitem[{Silva et~al.(2006)Silva, Frery, and Fatemi}]{glauber06}
\bibinfo{author}{G.~T. Silva}, \bibinfo{author}{A.~C. Frery},
  \bibinfo{author}{M.~Fatemi},
\newblock \bibinfo{title}{Image formation in vibro-acoustography with
  depth-of-fields effects},
\newblock \bibinfo{journal}{{C}omput. {M}ed. {I}maging {G}raph.}
  \bibinfo{volume}{30} (\bibinfo{year}{2006}) \bibinfo{pages}{321--327}.


\bibitem[{Jensen and Svendsen(1992)}]{jensen:262}
\bibinfo{author}{J.~A. Jensen}, \bibinfo{author}{N.~B. Svendsen},
\newblock \bibinfo{title}{Calculation of pressure fields from arbitrarily
  shaped, apodized, and excited ultrasound transducers},
\newblock \bibinfo{journal}{IEEE Trans. Ultrason. Ferroelec. Freq. Contr.}
  \bibinfo{volume}{39} (\bibinfo{year}{1992}) \bibinfo{pages}{262--267}.


\bibitem[{Yeoh and Zhang(2006)}]{yeoh2006}
\bibinfo{author}{W.-S. Yeoh}, \bibinfo{author}{C.~Zhang},
\newblock \bibinfo{title}{Constrained least squares filtering algorithm for
  ultrasound image deconvolution},
\newblock \bibinfo{journal}{IEEE Trans. Biomed. Eng.}
  \bibinfo{volume}{53} (\bibinfo{year}{2006}) \bibinfo{pages}{2001--2007}.
\bibitem[{Wang and Bovik(2002)}]{wang02}
\bibinfo{author}{Z.~Wang}, \bibinfo{author}{A.~C. Bovik},
\newblock \bibinfo{title}{A universal image quality index},
\newblock \bibinfo{journal}{{IEEE} {S}ignal {P}rocess {L}ett.}
  \bibinfo{volume}{9} (\bibinfo{year}{2002}) \bibinfo{pages}{81--84}.
\bibitem[{Wang et~al.(2002)Wang, Bovik, and Lu}]{WangBovikLu:ICASSP:02}
\bibinfo{author}{Z.~Wang}, \bibinfo{author}{A.~C. Bovik},
  \bibinfo{author}{L.~Lu},
\newblock \bibinfo{title}{Why is image quality assessment so
  difficult?},
\newblock in: \bibinfo{booktitle}{{P}roc. {IEEE} {I}nt. {C}onf. {A}coust. {S}peech
  {S}ignal {P}rocess}, volume~\bibinfo{volume}{4}, pp.
  \bibinfo{pages}{3313--3316} (\bibinfo{year}{2002}).

\bibitem[{Boutkedjirt and Reibold(2000) Boutkedjirt and Reibold}]{boutkedjirt:745}
\bibinfo{author}{T. Boutkedjirt}, \bibinfo{author}{R. Reibold},
\newblock \bibinfo{title}{Improvement of the lateral resolution
of finite-size hydrophones by deconvolution},
\newblock \bibinfo{journal}{Ultrasonics}
  \bibinfo{volume}{38} (\bibinfo{year}{2000}) \bibinfo{pages}{745--748}.

\bibitem[{Smith(1997)}]{smith97}
\bibinfo{author}{S.~W. Smith}, \bibinfo{title}{The Scientist and Engineer's
  Guide to Digital Signal Processing}, \bibinfo{publisher}{California Technical
  Pub.}, \bibinfo{year}{1997}, \bibinfo{address}{San Diego, CA, USA},
  p.~271.
\bibitem[{Wolfram(1999)}]{wolfram:hilbert}
\bibinfo{author}{E.~W. Weisstein}, \bibinfo{title}{``Hilbert Transform.'' from {MathWorld}-- a {Wolfram} web resource},
\bibinfo{year}{1999}, http://mathworld.wolfram.com/HilbertTransform.html. Last visit 09/01/2012. 
\end{thebibliography}
\end{document}